\documentclass[a4paper,fleqn]{cas-dc}

\usepackage[numbers,sort&compress]{natbib}

\usepackage{adjustbox}
\usepackage{float}
\usepackage[ruled]{algorithm2e}
\usepackage{algorithmic} 
\usepackage{CJKutf8}
\usepackage{enumitem}
\usepackage{multirow}
\usepackage[normalem]{ulem}
\useunder{\uline}{\ul}{}
\usepackage{amsmath}

\def\tsc#1{\csdef{#1}{\textsc{\lowercase{#1}}\xspace}}
\tsc{WGM}
\tsc{QE}
\tsc{EP}
\tsc{PMS}
\tsc{BEC}
\tsc{DE}

\usepackage{caption}
\begin{document}
\begin{sloppypar}
\let\printorcid\relax
\let\WriteBookmarks\relax
\def\floatpagepagefraction{1}
\def\textpagefraction{.001}
\begin{CJK}{UTF8}{gbsn}

\shorttitle{}

\shortauthors{J.Zhou et~al.}

\title [mode = title]{Recording Brain Activity While Listening to Music Using Wearable EEG Devices Combined with Bidirectional Long Short-Term Memory Networks}

\author[1]{Jingyi Wang}
\cormark[1]
\ead{xiabin126@126.com}

\author[2]{Zhiqun Wang}
\ead{hgzhou2020@163.com}

\author[3]{Guiran Liu}
\ead{gliu@sfsu.edu}

\affiliation[1]{organization={School of Music, Jiangxi Normal University},
    postcode={330027}, 
    city={Nanchang},
    country={China}}

\affiliation[2]{organization={School of Electronic Information, HuZhou College},
    postcode={313000}, 
    city={HuZhou},
    country={China}}

\affiliation[3]{organization={San Francisco State University},
    postcode={94132}, 
    city={San Francisco},
    country={United Stated}}

\cortext[cor1]{Corresponding author.}

\begin{abstract}
Electroencephalography (EEG) signals are crucial for investigating brain function and cognitive processes. This study aims to address the challenges of efficiently recording and analyzing high-dimensional EEG signals while listening to music to recognize emotional states. We propose a method combining Bidirectional Long Short-Term Memory (Bi-LSTM) networks with attention mechanisms for EEG signal processing. Using wearable EEG devices, we collected brain activity data from participants listening to music. The data was preprocessed, segmented, and Differential Entropy (DE) features were extracted. We then constructed and trained a Bi-LSTM model to enhance key feature extraction and improve emotion recognition accuracy. Experiments were conducted on the SEED and DEAP datasets. The Bi-LSTM-AttGW model achieved 98.28\% accuracy on the SEED dataset and 92.46\% on the DEAP dataset in multi-class emotion recognition tasks, significantly outperforming traditional models such as SVM and EEG-Net. This study demonstrates the effectiveness of combining Bi-LSTM with attention mechanisms, providing robust technical support for applications in brain-computer interfaces (BCI) and affective computing. Future work will focus on improving device design, incorporating multimodal data, and further enhancing emotion recognition accuracy, aiming to achieve practical applications in real-world scenarios.
\end{abstract}


\maketitle

\section{Introduction}  

The study of Electroencephalography (EEG) signals~\cite{hu2020scalp} has garnered significant attention in the fields of neuroscience and computer science. EEG signals, which reflect the brain's electrophysiological activity, are crucial tools for investigating brain function and cognitive processes. The advent of deep learning and wearable technology has revolutionized the ability to record and analyze EEG signals in real-time using portable devices~\cite{SXGY202402006}. This advancement not only facilitates research but also expands the applications in brain-computer interfaces (BCI) and affective computing~\cite{wang2022systematic,kamble2021ensemble}.

Music, as a complex auditory stimulus, profoundly influences emotional and cognitive functions of the brain~\cite{naser2021influence,daly2023neural}. Research indicates that different types of music can elicit various neural responses, thereby affecting emotional states. Recording brain activity via EEG while listening to music provides deep insights into the mechanisms by which music influences emotions. This is particularly valuable for applications in music therapy and other practical uses. However, the complexity and high dimensionality of EEG signals pose challenges in efficiently extracting and analyzing useful information~\cite{rahman2021recognition}.

This study addresses the critical need for advanced methods in EEG signal processing to enhance emotion recognition accuracy. Existing methods often struggle with the high dimensionality and complexity of EEG data. By introducing a novel combination of Bi-LSTM and attention mechanisms, this research aims to overcome these challenges and provide a robust solution for real-time emotion recognition using wearable EEG devices. The significance of this study lies in its potential applications in brain-computer interfaces, music therapy, and affective computing, where accurate emotion recognition can greatly enhance user experience and therapeutic outcomes.

Bidirectional Long Short-Term Memory (Bi-LSTM) networks~\cite{zheng2021attention,dai2024ai,wang2021machine,richardson2024reinforcement}, an advanced type of Recurrent Neural Network (RNN)~\cite{bouallegue2020dynamic,10438483,xu2022dpmpc,zhao2024task,song2024parallel}, are well-suited for this task. Bi-LSTM networks can leverage both past and future information in time series data, making them adept at capturing long-term dependencies. Applying Bi-LSTM to EEG signal analysis enhances the modeling capabilities for complex time-series data, thereby improving the accuracy and robustness of emotion recognition~\cite{GENG202438,zhang2024cunet,penglingcn,song2021efficient,wang2022classification}. Additionally, incorporating attention mechanisms allows Bi-LSTM to focus on critical features, further boosting model performance.

In this study, we utilize wearable devices to record EEG signals from participants while they listen to music. We then analyze these signals using a Bi-LSTM model to explore the impact of music on brain activity. The process involves low-pass filtering of EEG signals, feature extraction and selection, and the construction and training of the Bi-LSTM model. Our goal is to achieve efficient recording and precise prediction of brain activity. The results of this study will not only enhance the accuracy of emotion recognition but also provide substantial support for applications in BCI and affective computing.

This study addresses the critical need for advanced methods in EEG signal processing to enhance emotion recognition accuracy. Existing methods often struggle with the high dimensionality and complexity of EEG data. By introducing a novel combination of Bi-LSTM and attention mechanisms, this research aims to overcome these challenges and provide a robust solution for real-time emotion recognition using wearable EEG devices. The significance of this study lies in its potential applications in brain-computer interfaces, music therapy, and affective computing, where accurate emotion recognition can greatly enhance user experience and therapeutic outcomes.

The main contributions of our work are as follows:

\begin{enumerate}[label=(\arabic*)]
\item[$\bullet$] We propose a novel EEG signal processing method using Bi-LSTM and attention mechanisms, significantly enhancing emotion recognition accuracy.
\item[$\bullet$] Our method enables real-time brain activity recording and analysis under music stimulation using portable EEG devices.
\item[$\bullet$] The effectiveness of our model is validated on SEED and DEAP datasets, achieving a high accuracy of 98.28\% in emotion recognition.
\end{enumerate}

The remainder of this paper is structured as follows. Section~\ref{sec2} reviews related work on wearable EEG signal monitoring devices, EEG recording and MRI neuroimaging, applications of recurrent neural networks in EEG analysis, and auditory neural stimulation on the brain. Section~\ref{sec3} describes the proposed method, including EEG feature extraction, the construction of a 3D adjacency matrix of graph convolutional neural networks, and the Bi-LSTM model for EEG signal recognition. Section~\ref{sec4} presents relevant experimental results and analysis on the SEED and DEAP datasets. Finally, Section~\ref{sec5} concludes the paper with a summary and future research directions.

\section{Related Work}\label{sec2}

\subsection{Wearable EEG Signal Monitoring Devices}

Wearable EEG signal monitoring devices have gained widespread application in neuroscience, psychology, and biomedical engineering~\cite{zanetti2021real,srinivas2024support}. 
Their portability and ease of use make them essential tools in various domains, including affective computing, BCI, sleep management, emotion regulation, depression treatment, and fatigue monitoring. These devices, such as the Emotiv EPOC+ and Muse~\cite{kotowski2018validation}, enable real-time recording and analysis of an individual's emotional state by capturing EEG signals and identifying emotional changes, facilitating emotion regulation and psychological therapy~\cite{SXGY202306005}. In BCI technology, EEG devices decode brain signals to help users control external devices, such as enabling individuals with disabilities to operate wheelchairs. In sleep management, these devices analyze brain activity during sleep, providing feedback on sleep quality and promoting better sleep habits. They also aid in the long-term monitoring of emotional changes for diagnosing and treating emotional disorders, and in fatigue monitoring by providing real-time alerts during work or driving to prevent accidents.

Wearable EEG devices offer numerous advantages. Their portability and ease of use make them suitable for long-term wear and operation, meeting various application scenarios~\cite{shin2022wearable}. These devices provide real-time monitoring of brain signals, offering immediate feedback and monitoring results. Advanced devices like Mindeep support multiple electrode types, feature long battery life, impedance detection, and WiFi wireless transmission, and can provide high-quality raw data and multi-device synchronous data collection. However, these devices also have limitations. Despite offering higher sampling rates and data quality, many commercial devices still face constraints in sampling rate, signal resolution, and noise control. EEG signals are susceptible to artifacts from eye movements and muscle activity, necessitating complex preprocessing and signal processing algorithms, increasing data analysis complexity~\cite{li2022effect,de2022modeling,hong2024application,jin2023visual,dong2024design}. High-end devices are expensive, and cost considerations remain a factor for average users and some research institutions. Additionally, dry electrode devices may face issues with electrode contact and signal distortion, affecting data accuracy and stability~\cite{zhao2022preliminary,jin2024learning}.

Using these devices, researchers can efficiently record real-time brain activity data from participants while they listen to music, allowing for accurate analysis and prediction of brain responses to musical stimuli. Bi-LSTM models~\cite{,jiang2020dualvd,li2021evaluate,xie2023accel,chen2024enhancing,li2024utilizing} capture long-term dependencies in EEG signals and, combined with attention mechanisms, improve the extraction of key features. This method not only enhances the accuracy of emotion recognition but also provides robust technical support for BCI and affective computing applications. Future improvements in hardware design and signal processing algorithms can address current device limitations, further enhancing EEG signal processing efficiency and accuracy, offering broader applications in neuroscience research and healthcare.

\subsection{EEG Recording and MRI Neuroimaging}

Electroencephalography (EEG) and magnetic resonance imaging (MRI) are pivotal in neuroscience and biomedical engineering, each offering unique advantages~\cite{lambrecq2021association}. EEG involves placing electrodes on the scalp to capture the brain's electrical activity in real-time, boasting high temporal resolution. It is extensively used in affective computing, BCI, and cognitive neuroscience. Functional MRI (fMRI)~\cite{nentwich2020functional}, which measures blood oxygen level-dependent (BOLD) signals~\cite{palva2012infra}, provides high spatial resolution images of brain structure and function, making it invaluable for brain function localization, disease diagnosis, and cognitive research. Combining these techniques allows for multi-faceted brain activity analysis; for example, EEG can monitor real-time emotional changes while fMRI can precisely locate emotion-related brain activity.

Each method has distinct advantages and limitations. EEG's high temporal resolution enables millisecond-level recording of brain activity, ideal for studying rapid neural processes. Its portability and non-invasiveness make long-term monitoring and convenient operation possible, especially with wearable devices. The relatively low cost of EEG equipment makes it suitable for large-scale and daily monitoring. However, EEG's low spatial resolution limits its ability to reflect deep brain activity, and EEG signals are prone to artifacts, requiring complex preprocessing and signal processing algorithms. On the other hand, MRI's high spatial resolution provides detailed images of brain structures and functions. Its multi-modal imaging capabilities, such as combining structural imaging, functional imaging, and diffusion tensor imaging (DTI)~\cite{wilde2020diffusion}, offer comprehensive brain information. MRI is also non-invasive, suitable for repeated examinations. Nonetheless, MRI's low temporal resolution makes it challenging to capture rapid neural activity changes. The high cost and operational complexity of MRI equipment, requiring professional maintenance and operation, limit its widespread use. Additionally, MRI is sensitive to motion artifacts, necessitating strict control of the experimental environment.

\subsection{Applications of Recurrent Neural Networks in EEG Analysis}

Recurrent Neural Networks (RNNs)~\cite{tortora2020deep}, particularly Long Short-Term Memory (LSTM) and Bidirectional Long Short-Term Memory (Bi-LSTM) networks, are widely applied in EEG signal processing~\cite{teipel2009regional}. RNNs handle time series data, capturing temporal dependencies, making them particularly suitable for EEG signal analysis. LSTM networks address the vanishing gradient problem of traditional RNNs, making them ideal for long-sequence data analysis. Anitha and Hemanth proposed emotion recognition models based on LSTM and Gated Recurrent Unit (GRU) networks~\cite{9786857,li2024learning,RAN2024103664}, achieving efficient emotion classification through EEG signal processing, applicable in BCI and affective computing. Hybrid models like CNN-LSTM combine convolutional neural networks (CNN)~\cite{EINSHOKA2023399,zhu2024cross,li2024application,an2023runtime,liu2021investigation} with LSTM~\cite{sheykhivand2020recognizing,wang2024using,fellegara2023terrain,peng2023autorep,zhuang2020music} to extract spatial and temporal features, enhancing emotion recognition accuracy.

RNNs and their variants~\cite{,zhao2021investigation,de2023performance,liu2022preliminary} offer significant advantages and disadvantages in EEG signal processing. RNNs excel in handling dynamic time series data, making them ideal for analyzing rapidly changing EEG signals. LSTM networks effectively address the vanishing gradient problem of traditional RNNs, capturing long-term dependencies and improving the modeling of long-sequence data. Bi-LSTM networks further utilize both past and future information in sequences, enhancing the understanding of complex time series, showing excellent performance in emotion recognition and BCI systems. However, RNN models face challenges, including high computational complexity and long training times when processing large-scale data, requiring substantial computational resources. RNNs are also prone to overfitting, especially with high-dimensional data, necessitating regularization techniques and other methods to prevent overfitting. Despite LSTM networks mitigating the vanishing gradient problem, they may still encounter challenges when dealing with extremely long time sequences.

Using wearable EEG devices to record participants' brain activity in real-time while listening to music allows for accurate analysis and prediction of brain responses to musical stimuli. Bi-LSTM models capture long-term dependencies in EEG signals and, combined with attention mechanisms, improve the extraction of key features. This method not only enhances the accuracy of emotion recognition but also provides robust technical support for BCI and affective computing applications. Future advancements in RNN models and signal processing algorithms can better address current technical limitations, offering broader applications in neuroscience research and healthcare.

\subsection{Auditory Neural Stimulation on the Brain}

Auditory stimuli, particularly music, significantly impact the brain and are widely used in neuroscience and psychology research~\cite{somers2021eeg}. Music is known to activate multiple brain regions, including the auditory cortex, limbic system, and prefrontal cortex, thereby influencing emotional and cognitive functions. In affective computing, emotional responses induced by music are used to study the brain's emotion processing mechanisms~\cite{henao2020entrainment}. For instance, by analyzing EEG signals while listening to music, researchers can identify the emotional states elicited by different musical stimuli, which is valuable in music therapy and emotion computing. In cognitive neuroscience, music is used to study cognitive functions such as attention and memory. Additionally, auditory stimuli are applied in BCI systems to decode brain signals induced by auditory stimuli, enabling control of external devices.

Auditory stimulation offers numerous advantages in studying brain activity~\cite{ferster2022benchmarking}. Music, as a complex auditory stimulus, can activate multiple brain regions, providing researchers with a comprehensive understanding of the brain's emotional and cognitive processing mechanisms. Music stimuli are easy to control and standardize, ensuring high repeatability and reliability of experimental results. Furthermore, the effects of music on the brain are apparent and significant, allowing researchers to clearly observe changes in brain activity through EEG signals~\cite{GENG20224807}.

However, auditory stimulation also presents some challenges. Firstly, individual differences in emotional responses to music can significantly affect the generalizability and consistency of experimental results. Secondly, music-induced EEG signals are complex, involving multiple frequency bands and brain regions, requiring sophisticated signal processing and analysis techniques. Additionally, the effects of music on the brain are transient, posing challenges in effectively capturing and analyzing these brief changes.

By using wearable EEG devices to record brain activity while participants listen to music, researchers can monitor and analyze the brain's response to musical stimuli in real-time. Combining these recordings with Bi-LSTM networks allows for more accurate analysis and prediction of the brain's emotional and cognitive responses to different musical stimuli. Bi-LSTM models capture long-term dependencies in EEG signals and, combined with attention mechanisms, improve the extraction of key features. This combined approach not only enhances emotion recognition accuracy but also provides strong technical support for BCI and affective computing applications. Future research should focus on optimizing auditory stimulus control and EEG signal analysis methods to better understand and utilize the effects of auditory neural stimulation on the brain, providing broader applications in neuroscience and psychology.

\section{Method}\label{sec3}

\begin{figure*}[h]
	\centering
	\includegraphics[width=1.0\textwidth]{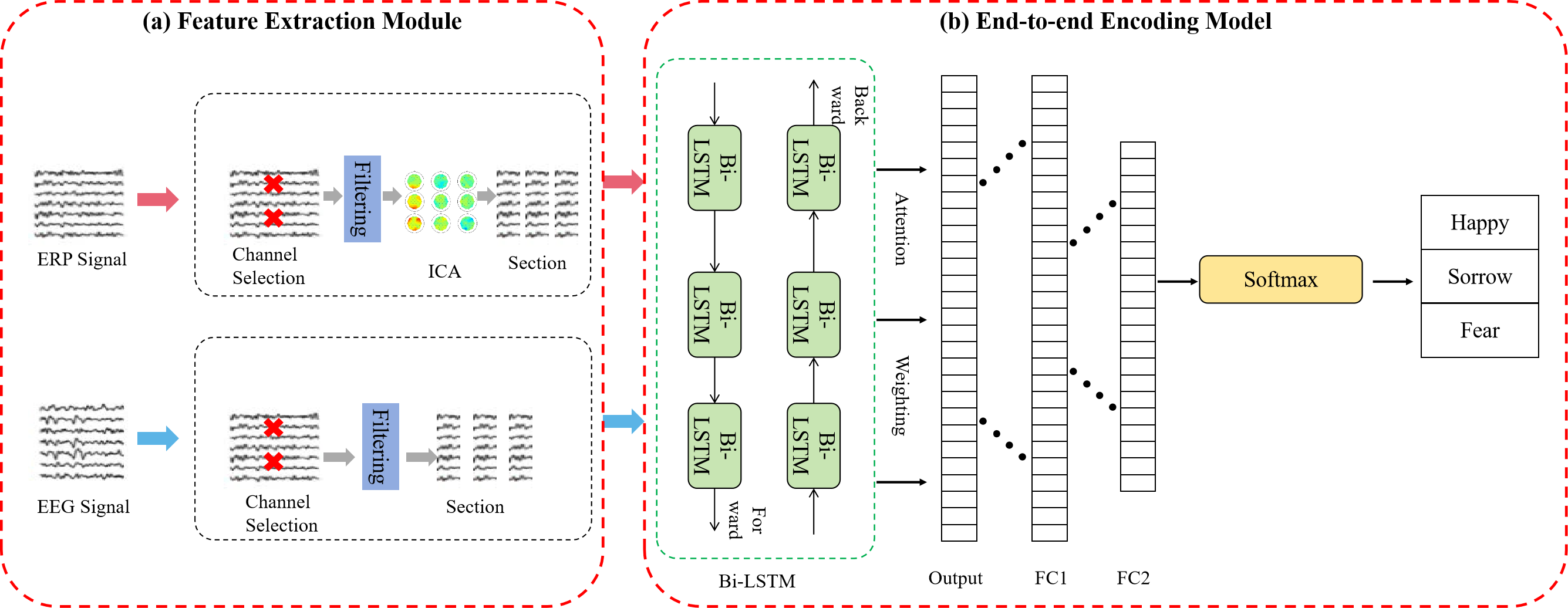}
	\caption{Schematic Diagram of the Proposed Bi-LSTM Framework for Recording Brain Activity While Listening to Music. EEG Signals Are Extracted from the SEED Dataset, and ERP Signals Are Extracted from the DEAP Dataset. The Two Datasets Are Processed Independently and Do Not Interfere with Each Other.}
	\label{f1}
\end{figure*}

As shown in Figure~\ref{f1}, the overall experimental model involves extracting features from the EEG and ERP signals of the SEED and DEAP datasets. We processed these signals in segments before applying the Bi-LSTM model~\cite{deng2024solving,patel2022inspection,lee2024traffic,zhu2019image,liu2022impact}. The Bi-LSTM architecture comprises an enhanced Bi-LSTM layer, an attention weighting layer, two fully attentive layers, and a Softmax classification layer. For feature extraction of the P300 component in ERP data and general EEG signals, we performed channel selection, filtering, and segmentation. However, the extraction of P300 features from ERP data included an additional step of Independent Component Analysis (ICA) to remove prominent signal noise and artifacts.

The subsequent Bi-LSTM model incorporated our attention gate method, applying attention weighting to both types of signal features. We utilized dropout techniques in the fully connected layers to prevent overfitting. Separate experiments were conducted on the SEED and DEAP datasets. The SEED dataset experiments primarily focused on processing EEG signals, while the DEAP dataset experiments concentrated on ERP signal processing.

\subsection{EEG Feature Extraction}

In the experiment, the differential entropy (DE)~\cite{zhang2020automatic} feature in the frequency domain is used as the input for emotion recognition. The extraction process of the differential entropy (DE) feature in the frequency domain is as follows.

Differential entropy (DE) extends the Shannon information entropy \(H(X)=-\sum_{x}p(x)\log\bigl(p(x)\bigr)\) to continuous random variables, as shown in Equation~\ref{equation1}.

\begin{equation}
DE=-\int_{a}^{b}p(x)\log\bigl(p(x)\bigr)dx,
\label{equation1}
\end{equation}

where \(p(x)\) represents the continuous probability density function, and \([a,b]\) represents the interval of information extraction. For an EEG signal segment that approximately follows a Gaussian distribution \(N(\mu,\sigma^{2})\), its differential entropy is equal to the logarithm of its energy spectrum in a specific frequency band, as shown in Equation~\ref{equation2}.

\begin{equation}
\begin{aligned}
DE& =-\int_{a}^{b}\frac{1}{\sqrt{2\pi\sigma^{2}}}e^{-\frac{(x-\mu)^{2}}{2\sigma^{2}}}\log{(\frac{1}{\sqrt{2\pi\sigma^{2}}}e^{-\frac{(x-\mu)^{2}}{2\sigma^{2}}})}dx  \\
&=\frac{1}{2}\log{(2\pi e\sigma^{2})},
\end{aligned}
\label{equation2}
\end{equation}

\subsection{Construction of 3D Adjacency Matrix of Graph Convolutional Neural Networks}

In EEG-based emotion recognition research utilizing graph neural networks~\cite{zhou2024adapi,peng2024maxk,jin2024apeer}, it is known that describing the relationships between different EEG electrode channels, i.e., constructing an adjacency matrix, is crucial for EEG emotion classification. In this section, we use the spatial distance between EEG nodes to construct the adjacency matrix representing the topology of EEG channels. The specific construction process is as follows:

The adjacency matrix $\mathrm{A} \in \mathbb{R}^{n \times n}$, where $n$ denotes the number of channels in the EEG signals. Each entry $a_{ij}$ is learnable and represents the connection weight between channel $i$ and channel $j$. The international 10-20 EEG electrode system provides the three-dimensional coordinates of each electrode mapped onto a unit sphere. The physical distance between two electrodes is used to measure the connection relationship in the brain space. The farther the distance, the less tightly the channels are connected. Suppose the coordinates of two points on the sphere with radius $r$ are $(x_i, y_i, z_i)$ and $(x_j, y_j, z_j)$, the distance $d_{ij}$ between the two points in Cartesian space can be expressed as shown in Equation~\ref{equation3}.

\begin{equation}
d_{ij}=\arccos\left(\frac{x_ix_j + y_iy_j + z_iz_j}{r^2}\right),
\label{equation3}
\end{equation}

\begin{figure}[h]
	\centering
	\includegraphics[width=0.5\textwidth]{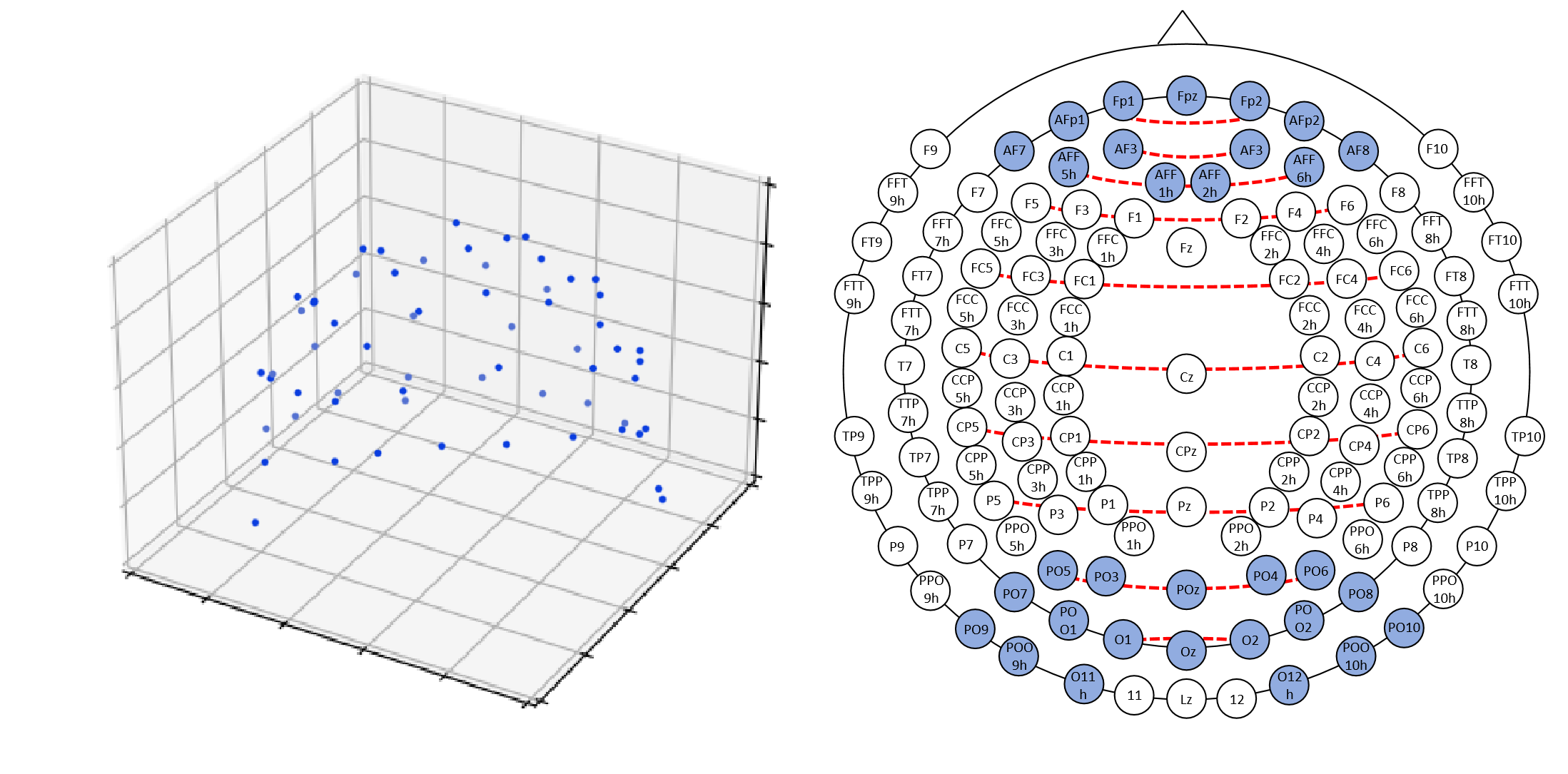}
	\caption{3D location map of EEG channels and location channel connectivity map. Left: 3D location map of EEG channels; Right: Initialize the global channel connectivity graph.}
	\label{f2}
\end{figure}

Figure~\ref{f2} below is a schematic diagram of the 3D spatial relationship of EEG channels used to construct the model's adjacency matrix. The points in the 3D graph represent the electrode positions of the wearable EEG device used for measuring brainwave activity.

In sparse fMRI networks, using about 20\% of all possible connections typically maximizes the efficiency of network topology. Therefore, for each EEG channel, we retain connections to its nearest $K$ channels, considering them as connected. The value of $K$ is chosen based on the number of electrode channels used in the EEG data acquisition equipment. For devices with 62 electrode channels, the value of $K$ is selected as $62 \times 20\% \approx 12$.

The asymmetry of neural activity between the left and right hemispheres is significant in valence and arousal prediction. Therefore, we select certain electrode pairs to initialize the adjacency matrix. To leverage the asymmetry information between the left and right hemispheres, we use 9 global connection pairs and initialize the global inter-channel relationships in the adjacency matrix as shown in Equation~\ref{equation4}:

\begin{equation}
a_{ij}=a_{ij}+1,
\label{equation4}
\end{equation}

As shown in Figure 3, the global channel pairs $(i,j)$ used for initialization are (FP1, FP2), (AF3, AF4), (F5, F6), (FC5, FC6), (C5, C6), (CP5, CP6), (P5, P6), (PO5, PO6), and (O1, O2).

\subsection{Bi-LSTM constructs EEG signal recognition}

Bi-LSTM is a deep learning model based on LSTM. LSTM networks, through their unique gating mechanisms, can effectively capture and learn long-term dependencies in sequential data, addressing the issues of gradient vanishing and exploding in traditional RNNs (Recurrent Neural Networks).

\begin{figure}[h]
	\centering
	\includegraphics[width=0.5\textwidth]{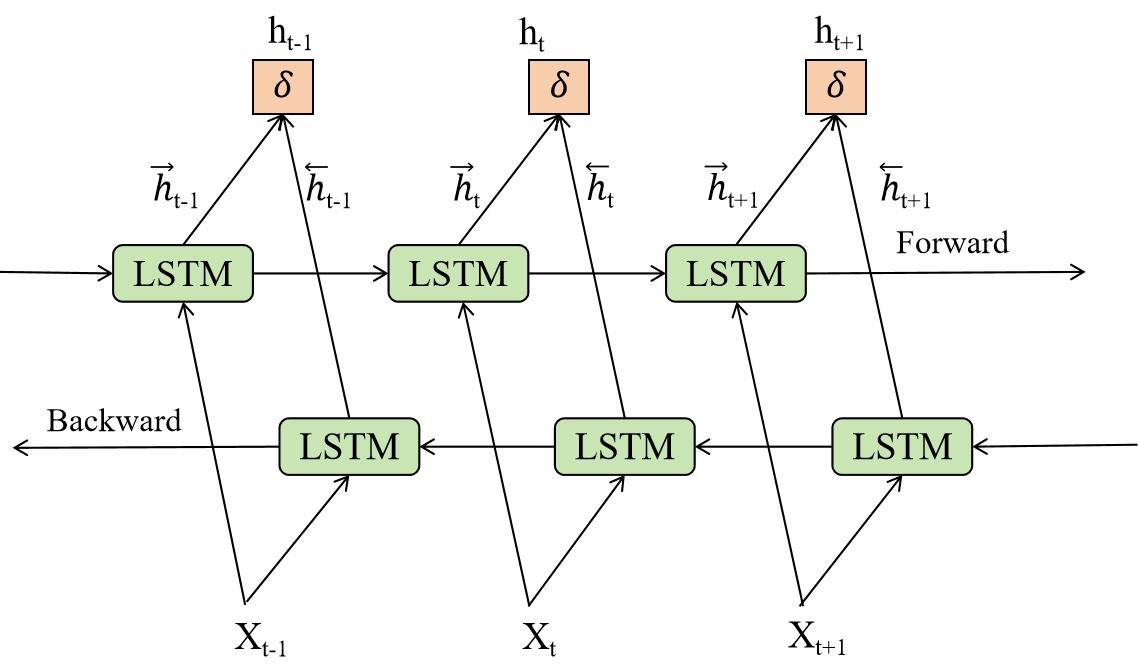}
	\caption{Bidirectional Long Short-Term Memory Network Network}
	\label{f3}
\end{figure}

As shown in Figure~\ref{f3}, unlike traditional unidirectional LSTM, Bi-LSTM comprises two LSTM layers: one processes the sequence forward, and the other processes it backward. This bidirectional mechanism enables Bi-LSTM to leverage both past and future information in the sequence, enhancing the model's understanding of time-series data.

EEG (Electroencephalography) signals are complex time-series data containing rich temporal and frequency information. Due to the dynamic nature of EEG signals, traditional signal processing methods struggle to capture long-term dependencies and dynamic changes effectively. Bi-LSTM, with its bidirectional mechanism, can more accurately model the temporal information in EEG signals, improving the accuracy of emotion recognition.

In EEG signal recognition models, Bi-LSTM processes EEG signals through forward and backward LSTM layers to extract deep features from the signals. These features can be used for further classification or regression tasks, aiding in recognizing participants' brain activity patterns in different emotional states. By incorporating attention mechanisms, the Bi-LSTM model can further focus on important moments in the signal, enhancing the model's recognition capabilities.

In Bi-LSTM, the forward LSTM layer processes EEG samples from time index 1 to \( t \), generating the forward hidden state sequence \( \overrightarrow{h_t} \), while the backward LSTM layer processes EEG samples from time index \( t+1 \) to the end, generating the backward hidden state output \( \overleftarrow{h_t} \). This bidirectional processing mechanism allows Bi-LSTM to capture dynamic changes in EEG signals and more accurately model long-term dependencies in the signals.

\begin{figure*}[h]
	\centering
	\includegraphics[width=1.0\textwidth]{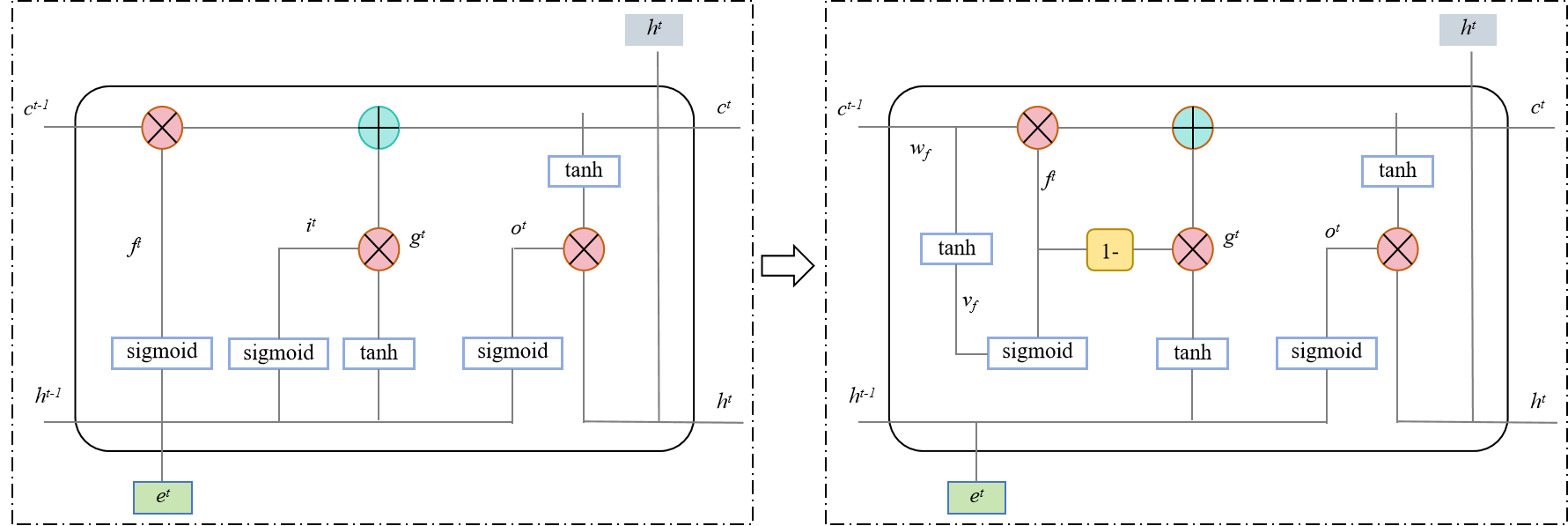}
	\caption{Single LSTM Cell Diagram. Left: Original LSTM Cell Diagram; Right: LSTM Cell Diagram with Attention Gate.}
	\label{f4}
\end{figure*}

The left part of Figure~\ref{f4} illustrates the original architecture of LSTM, where the cell's output state update is related to the previous hidden layer output and the current input. At each time step \( t \), the LSTM unit receives the current input \( x_t \), the hidden state \( h_{t-1} \), and the cell state \( c_{t-1} \) from the previous time step. The LSTM controls information storage and transfer through three gating mechanisms: the input gate, the forget gate, and the output gate. The input gate \( i_t \) determines the update of the cell state by the current input, the forget gate \( f_t \) decides whether to retain the previous cell state \( c_{t-1} \), and the output gate \( o_t \) decides whether to transfer the hidden state \( h_{t-1} \) to the next LSTM unit. The candidate cell state \( \tilde{c_t} \) is a nonlinear transformation of the current input and the previous hidden state. The current cell state \( c_t \) is updated through the forget gate and the input gate, and the current hidden state \( h_t \) is determined by the output gate and the nonlinear transformation of the current cell state. Here, \( \sigma \) represents the sigmoid activation function, \( \tanh \) represents the tanh activation function, \( \odot \) denotes element-wise multiplication, \( W \) and \( U \) are weight matrices, and \( b \) is the bias vector. The formulas for the LSTM unit are shown below:

\begin{equation}
\begin{aligned}
i_t &= \sigma(W_i x_t + U_i h_{t-1} + b_i), \\
f_t &= \sigma(W_f x_t + U_f h_{t-1} + b_f), \\
o_t &= \sigma(W_o x_t + U_o h_{t-1} + b_o), \\
\tilde{c_t} &= \tanh(W_c x_t + U_c h_{t-1} + b_c), \\
c_t &= f_t \odot c_{t-1} + i_t \odot \tilde{c_t}, \\
h_t &= o_t \odot \tanh(c_t),
\end{aligned}
\label{equation5}
\end{equation}

Bi-LSTM combines the outputs of the forward and backward LSTM layers to form the final hidden state output \( h_t \). The forward hidden state is computed by the forward LSTM layer, and the backward hidden state is computed by the backward LSTM layer. By combining the forward and backward hidden states, Bi-LSTM can more comprehensively understand the temporal information in EEG signals, leading to higher accuracy in emotion recognition. The formulas for the Bi-LSTM unit are shown below:

\begin{equation}
\begin{aligned}
\overrightarrow{h_t} &= \text{LSTM}_{\rightarrow}(x_t, \overrightarrow{h_{t-1}}), \\
\overleftarrow{h_t} &= \text{LSTM}_{\leftarrow}(x_t, \overleftarrow{h_{t+1}}), \\
h_t &= [\overrightarrow{h_t}; \overleftarrow{h_t}],
\end{aligned}
\label{equation6}
\end{equation}

In emotion recognition models, Bi-LSTM effectively captures the temporal dynamics in EEG signals through its bidirectional mechanism. Using wearable frontal EEG signal monitoring devices, EEG signals of participants are recorded while they listen to different types of music. The device employs dry electrode technology, simplifying the signal acquisition process and making it suitable for large-scale daily applications. The collected EEG signals undergo preprocessing, including denoising, filtering, and feature extraction, to ensure signal quality and analysis accuracy. The Bi-LSTM model processes the preprocessed EEG signals through forward and backward LSTM layers, extracting deep features. By incorporating attention mechanisms to focus on important moments in the signals, this study introduces an improved LSTM method, as shown on the right side of Figure~\ref{f4}, which incorporates the attention mechanism to capture essential historical information and update the cell state. This ultimately improves the accuracy and robustness of emotion recognition, as represented by Eq.~\ref{equation7}:

\begin{equation}
f_t = \delta(v_f * \tanh(w_f * c_{t-1})).
\label{equation7}
\end{equation}

where \( v_f \) and \( w_f \) are the parameters of the attention mechanism. This method reduces the dimensionality of the training parameters compared to Eq.~\ref{equation5}.

\section{Experiment}\label{sec4}

\subsection{Datasets}

\textbf{SEED dataset.} The SEED dataset~\cite{cimtay2020investigating}, collected by Shanghai Jiao Tong University, involves participants watching movie clips of various emotional types, with their EEG signals recorded in response to these stimuli. This dataset was collected using a 62-channel EEG cap. The experiment included 15 participants, comprising 8 males and 7 females. Each participant was required to attend three identical sessions, with each session spaced 7-14 days apart. In each session, participants watched 15 movie clips, as detailed in Table~\ref{data1}. These 15 clips were selected from different segments of 6 movies, with each segment eliciting different emotions and affective states. During the experiment, the length of each clip was limited to approximately 4 minutes. After watching each clip, participants completed an emotional assessment and took a rest. Once the EEG data was acquired, it was downsampled and filtered for further analysis.

\begin{table}[]
\centering
\caption{SEED dataset composition.}
\begin{tabular}{l|l|l}
\hline
Number & Name of the clip           & Label    \\ \hline
1      & Lost in Thailand           & Vigorous \\
2      & World Heritage in China    & Neutral  \\
3      & Aftershock                 & Passive  \\
4      & Back to 1942               & Passive  \\
5      & Flirting Scholar           & Vigorous \\
6      & Just Another Pandora's Box & Vigorous \\ \hline
\end{tabular}
\label{data1}
\end{table}

\textbf{DEAP dataset.} The DEAP dataset~\cite{khateeb2021multi}is a multimodal dataset specifically designed for affective analysis, offering a rich collection of electroencephalogram (EEG), physiological signals, and video data. This dataset involves 32 participants who watched 40 one-minute-long music video excerpts, during which their emotional responses were recorded. The first part of the dataset comprises evaluations of the music videos by 14 to 16 participants, who rated the videos based on arousal, valence, and dominance. The second part of the dataset includes ratings, physiological recordings, and facial videos from 32 volunteers while watching the aforementioned 40 music video excerpts. In addition to emotional state ratings, the physiological recordings primarily consist of EEG data. The objective of the DEAP dataset is to provide researchers with a standardized dataset for testing and validating their methods of estimating emotional states, thereby advancing research in affective computing, emotion recognition, and brain-computer interfaces.

\subsection{Data Preprocessing and Feature Extraction}

EEG signals have low amplitudes and are often contaminated with various noise signals during acquisition. The preprocessing process aims to enhance the quality of EEG signals and reduce noise and artifacts. Common artifacts include electrooculographic (EOG) artifacts, electromyographic (EMG) artifacts, electrocardiographic (ECG) artifacts, skin conductance responses, and power line interference.

The EEG preprocessing workflow includes the following steps:
\begin{enumerate}[label=(\arabic*)]
\item Downsampling: The original EEG signals were downsampled to reduce computational complexity (SEED: 1000 Hz to 200 Hz; DEAP: 512 Hz to 128 Hz).
\item Bad channel detection and removal: Channels with excessive noise or artifacts were identified and removed.
\item Electrode re-referencing: Signals were re-referenced to a common average reference.
\item Bandpass filtering: A bandpass filter (0.5-50 Hz) was applied to remove irrelevant frequencies and noise.
\item Artifact removal using ICA: Independent Component Analysis (ICA) was performed to remove EOG, EMG, and other artifacts.
\end{enumerate}

\subsection{Emotion Evoked EEG Signal Generation and Acquisition}

EEG is a method of recording the electrophysiological activity of the brain's neural tissues on the surface of the cerebral cortex. Neuronal excitation and inhibition in the brain generate voltage fluctuations, typically measured from the scalp in the range of 10 to 100 $\mu$V in adults. Information in EEG signals is primarily contained within the frequency spectrum of 0.5 to several tens of Hertz. Based on frequency differences, EEG signals can be classified into five bands: $\delta$ band (Delta, 1-4 $Hz$), $\theta$ band (Theta, 4-8 $Hz$), $\alpha$ band (Alpha, 8-12 $Hz$), $\beta$ band (Beta, 12-30 $Hz$), and $\gamma$ band (Gamma, $\textgreater$ 30 Hz). The rhythmic characteristics of each band are detailed in Table~\ref{tab1}.

\begin{table}[]
\centering
\caption{The rhythm of each band of EEG signal.}
\resizebox{\linewidth}{!}{
\begin{tabular}{c|c|c}
\hline
\textbf{Band}    & \textbf{Frequency ($Hz$)} & \textbf{Human State}                                                                        \\ \hline
Delta $(\delta)$ & 0.1-3.0                 & \begin{tabular}[c]{@{}c@{}}Deep sleep, disordered, \\ hypoxic, comatose states\end{tabular} \\ \hline
Theta $(\theta)$ & 4.0-7.0                 & \begin{tabular}[c]{@{}c@{}}Fatigue, depression, \\ low mood, disappointment\end{tabular}    \\ \hline
Alpha $(\alpha)$ & 8.0-12.0                & \begin{tabular}[c]{@{}c@{}}Relaxed, calm, eyes \\ closed but awake\end{tabular}             \\ \hline
Beta $(\beta)$   & 12.5-28.0               & Tense, excited,happy                                                                        \\ \hline
Gamma $(\gamma)$ & 29.0-50.0               & \begin{tabular}[c]{@{}c@{}}Highly aroused, \\ excited, tense\end{tabular}                   \\ \hline
\end{tabular}
}
\label{tab1}
\end{table}

\begin{figure}[h]
	\centering
	\includegraphics[width=0.5\textwidth]{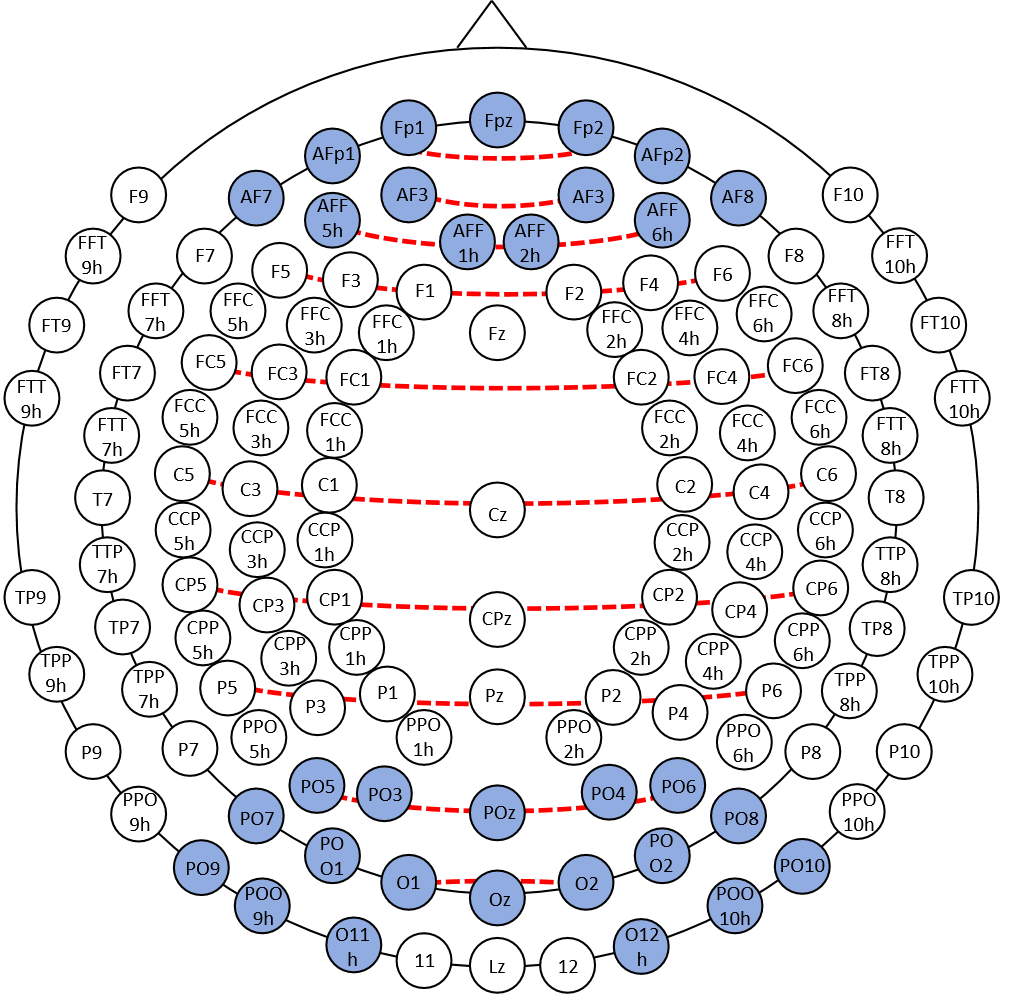}
	\caption{EEG electrode 130 system placement method.  The blue labels indicate the 31 EEG channels selected for the second phase of the experiment, with 13 channels in the prefrontal cortex and 18 channels in the occipital lobe.}
	\label{f5}
\end{figure}

EEG signal collection generally employs either dry electrode or wet electrode methods. The current predominant method for EEG data collection is the wet electrode method, which has the advantage of obtaining more distinct EEG data but is inconvenient and less suitable for practical, everyday collection. The use of dry electrodes for EEG signal collection does not require an electrolyte, allowing the electrodes to contact the scalp directly, thereby offering greater convenience. However, due to the higher impedance of the stratum corneum, the EEG signals collected in this manner tend to be weaker. Figure~\ref{f5} illustrates the layout of the 130-electrode system used in this study, which is both easy to implement and ensures test reproducibility. In the second phase of the experiment, 31 EEG channels were selected (indicated by the blue electrodes in Figure~\ref{f5}), with 13 channels located in the prefrontal cortex and 18 channels in the occipital lobe.

EEG signals have low amplitudes and are often contaminated with various noise signals during acquisition. The preprocessing process aims to enhance the quality of EEG signals and reduce noise and artifacts. Common artifacts include electrooculographic (EOG) artifacts, electromyographic (EMG) artifacts, electrocardiographic (ECG) artifacts, skin conductance responses, and power line interference.

EOG artifacts typically occur in the frequency range of 0-15 Hz and are common EEG artifacts usually associated with eye movements. Blinking artifacts are characterized by narrow spikes with large amplitudes. EMG artifacts generally have higher frequencies and are mixed into EEG signals due to muscle movements during EEG experiments. ECG artifacts primarily arise from the signals generated by myocardial contraction and relaxation, typically presenting as low-frequency signals (less than 5 Hz). Skin conductance responses usually appear in signals collected from the palms or fingertips. Power line interference originates from the 50 Hz mains electricity voltage.

The EEG preprocessing workflow includes downsampling, bad channel detection, electrode re-referencing, bandpass filtering, and Independent Component Analysis (ICA)~\cite{antony2022classification} for artifact removal.

\subsection{Bi-LSTM Construction and Simulation Parameters}

The Bi-LSTM model employed in this study is designed to process EEG signals effectively, leveraging both past and future information to capture long-term dependencies and enhance emotion recognition accuracy. Below are the details of the model construction, hyperparameters, time settings, and experimental parameters:

\textbf{Bi-LSTM Model Architecture.} The Bi-LSTM architecture consists of the following components:
An enhanced Bi-LSTM layer with 128 hidden units in each direction (forward and backward); An attention weighting layer to emphasize critical features in the EEG signals; Two fully connected layers with 64 neurons each, using ReLU activation functions; A Softmax classification layer to output the probability distribution of emotional states.

\textbf{Hyperparameters.} The hyperparameters for the Bi-LSTM model were carefully selected based on empirical experiments and existing literature. The following settings were used:
Learning rate: 0.001;Batch size: 64; Epochs: 100; Dropout rate: 0.5 to prevent overfitting; Optimizer: Adam.

\textbf{Time Settings and Data Preprocessing.} For the SEED dataset, EEG signals were recorded at a sampling rate of 1000 Hz and then downsampled to 200 Hz to reduce computational complexity. Each recording session lasted approximately 4 minutes per clip. The EEG signals were segmented into 1-second non-overlapping windows, resulting in 200 data points per segment.

For the DEAP dataset, EEG signals were recorded at a sampling rate of 512 Hz and then downsampled to 128 Hz. Each music video excerpt lasted 1 minute, and the signals were segmented into 1-second windows with 128 data points per segment.

\textbf{Experimental Parameters and Settings.} The experimental setup included the following key parameters:
Feature extraction: Differential Entropy (DE) features were extracted from the segmented EEG signals; Model training: The Bi-LSTM model was trained on 80\% of the data and validated on the remaining 20\%; Evaluation metrics: Accuracy, precision, recall, and F1-score were used to evaluate the model's performance.

\textbf{Simulation Process.}The simulation of EEG signal fluctuations was conducted as follows:
The Data\_preprocessed files from the DEAP dataset were fed into a linear filter to map the spectral output of the EEG signals, representing brain neural activity; A receptive field model of the linear filter was created to generate artificial neural responses; The data were downsampled along the time dimension, and the receptive field was used to create an artificial neural response by performing a dot product with the receptive field; The Bi-LSTM model was then used to simulate and predict the time-varying stimulus response of the EEG signals.

\subsection{Results}

To ensure a fair comparison between the different models, we maintained consistent experimental conditions across all tests. All models were trained and evaluated using the same datasets (SEED and DEAP) and preprocessing techniques. Additionally, we used the same training, validation, and testing splits to provide an equitable basis for performance evaluation. Hyperparameters for each model were tuned individually to achieve optimal performance, ensuring that each model was fairly optimized for comparison. We also employed standard evaluation metrics, such as accuracy, precision, recall, and F1-score, to provide a comprehensive assessment of each model's performance. The results demonstrate the superior performance of the Bi-LSTM model with attention mechanisms, highlighting the effectiveness of our proposed approach in emotion recognition using EEG signals.

\textbf{Results on SEED dataset.} We conducted a comparative experiment using EEG-based object detection visual models on the SEED dataset. The models we employed include Support Vector Machine (SVM)~\cite{kang2020identification}, Pyramid Match Kernel (PMK)~\cite{hou2023eeg}, EEG-Net~\cite{huang2022virtual}, LSTM~\cite{kadri2023new}, and Bi-LSTM. SVM is a supervised learning model used for classification and regression analysis, effectively handling high-dimensional data. PMK is a kernel function used for image matching by processing image features through a pyramid structure to achieve efficient matching. EEG-Net is a deep learning model specifically designed for processing and analyzing EEG data to improve the classification and recognition performance of EEG signals. LSTM is a type of RNN model that addresses long-term dependency issues through its gating mechanism, making it suitable for handling sequential data. Bi-LSTM consists of forward and backward LSTM layers, enabling the model to utilize both past and future information in the sequence, thus enhancing its understanding of temporal data.

Our proposal and the performance comparison of these models are shown in Table~\ref{tab2}. In addition to these advanced model algorithms, we also compared our model with the incorporation of attention gates and attention weights to explore the effectiveness of Bi-LSTM. As shown in the table, the Bi-LSTM models with attention mechanisms significantly improved accuracy. Specifically, the Bi-LSTM-AttGW (1 layer and 128 hidden size) model achieved the highest accuracy of 98.28\% in multiclass tasks. This demonstrates that incorporating attention mechanisms can significantly enhance the performance of models in emotion recognition using EEG signals. This method greatly improves the accuracy of capturing EEG signals.

\begin{table*}[]
\caption{Comparative Performance Analysis of Advanced Models on the SEED Dataset for EEG-Based Emotion Recognition.}
\begin{tabular}{llll}
\hline
Model                                               & Multiclass         & Channel      & Accuracy(\%)   \\ \hline
SVM~\cite{kang2020identification}                                                 & Binary             & 256          & 82.52          \\
EEG-Net~\cite{huang2022virtual}                                             & Binary             & 14           & 88.04          \\
PMK~\cite{hou2023eeg}                                                 & Multi(3)           & 32           & 91.45          \\
LSTM~\cite{kadri2023new}                                                & Multi(20)          & 64          & 86.75          \\
Bi-LSTM                                             & Multi(20)          & 64          & 92.39          \\
Bi-LSTM+AttW(1 layer and 64 hidden size)           & Multi(20)          & 64          & 93.42          \\
Bi-LSTM+AttG(1 layer and 64 hidden size)           & Multi(20)          & 64         & 95.16          \\
Bi-LSTM+AttWG(1 layer and 32 hidden size)           & Multi(20)          & 64          & 95.08          \\
\textbf{Bi-LSTM+AttWG(1 layer and 64 hidden size)} & \textbf{Multi(20)} & \textbf{128} & \textbf{98.28} \\
Bi-LSTM+AttWG(1 layer and 128 hidden size)          & Multi(20)          & 64          & 96.22          \\ \hline
\end{tabular}
\label{tab2}
\end{table*}

As shown in Table~\ref{tab2b}, we compared the accuracy and standard deviation of different models in various frequency bands of the SEED dataset. Our proposed model demonstrated superior performance across all frequency bands, particularly in the $\gamma$ band, where it achieved an accuracy of 98.28\% with a standard deviation of 7.42. This highlights the robustness and effectiveness of our Bi-LSTM model with attention mechanisms in capturing the complexities of EEG signals and improving emotion recognition accuracy.

\begin{table}[]
\caption{Comparison of Accuracy and Standard Deviation of Different Models in Different Frequency Bands of SEED Dataset.}
\resizebox{\linewidth}{!}{
\begin{tabular}{cccc}
\hline
\begin{tabular}[c]{@{}c@{}}Frequency \\ Band\end{tabular} & SVM~\cite{kang2020identification}               & LSTM~\cite{kadri2023new}                       & Ours                       \\ \hline
$\delta$                                                & 60.12 $\pm$ 14.11 & \textbf{74.45 $\pm$ 11.43} & 70.26 $\pm$ 15.36          \\
$\theta$                                                & 60.92 $\pm$ 10.25 & 71.26 $\pm$ 5.46           & \textbf{75.34 $\pm$ 15.16} \\
$\alpha$                                                & 66.95 pm 14.85    & 74.31 $\pm$ 12.58          & \textbf{86.13 $\pm$ 14.83} \\
$\beta$                                                 & 80.71$\pm$ 11.16  & 81.25 $\pm$ 10.11          & \textbf{90.23 $\pm$ 9.13}  \\
$\gamma$                                                & 82.52 $\pm$ 9.17  & 86.75 $\pm$ 8.43           & \textbf{98.28 $\pm$ 7.42}  \\ \hline
\end{tabular}
}
\label{tab2b}
\end{table}

\begin{figure}[h]
	\centering
	\includegraphics[width=0.5\textwidth]{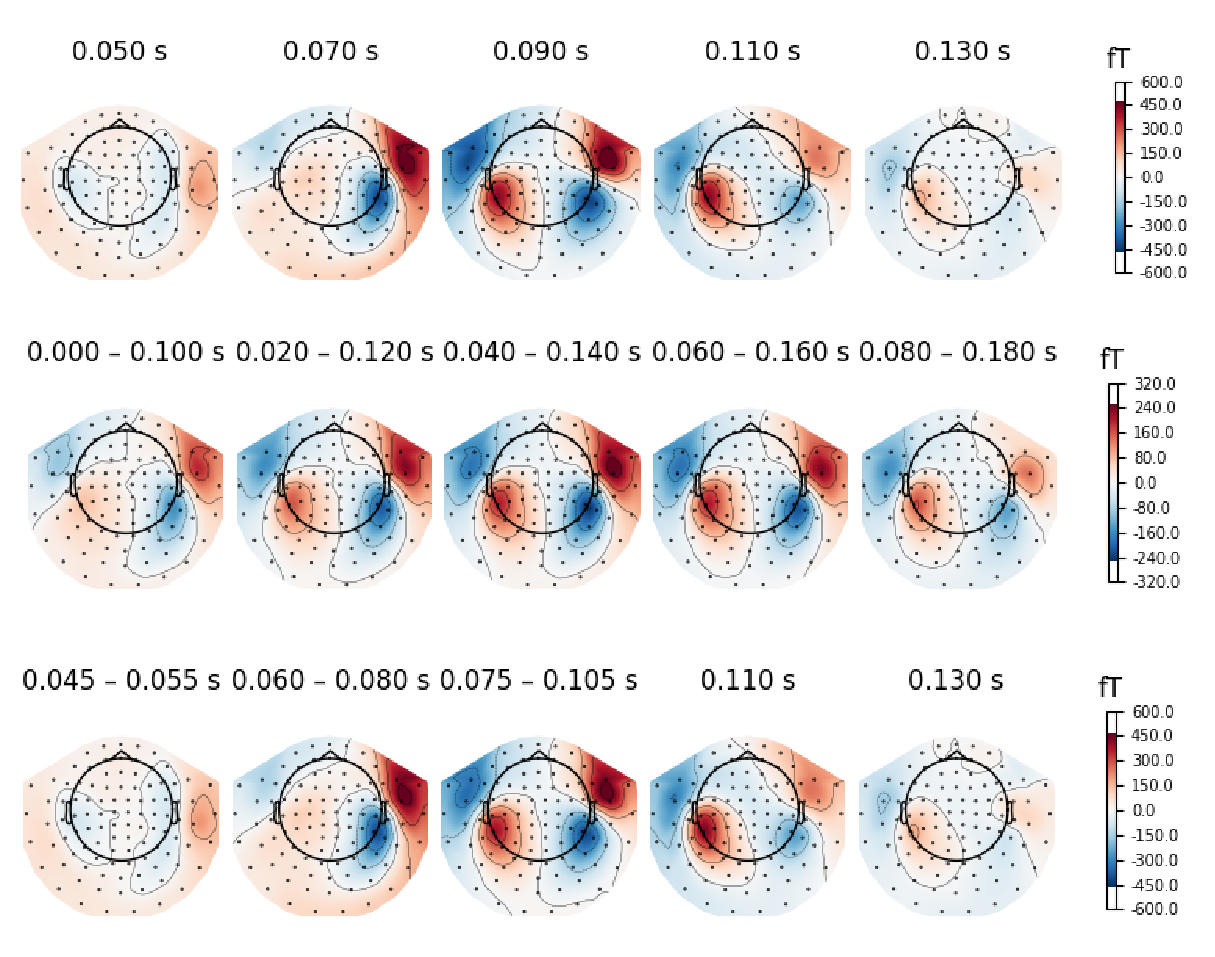}
	\caption{Dynamic Changes in EEG Signal Regions of Interest at Different Time Points Across Various Movie Scenes in the SEED Datazset. The figure illustrates how different emotional stimuli from movies affect brain activity recorded through EEG signals.}
	\label{f6}
\end{figure}

To visualize the EEG signal activity of the brain, we applied a low-pass filter at 8 $Hz$ to the EEG signals from the SEED dataset to remove high-frequency noise. Figure~\ref{f6} shows the EEG signal fragments in the regions of interest (ROIs) using the 130-electrode system layout. Each row in the figure represents the brain activity recorded for a segment from different movies, illustrating the variations in EEG signals in response to different emotional stimuli.

The first row represents the brain activity recorded for a segment from the movie \textit{Lost in Thailand}, showing EEG signal snippets from 50 ms before to 130 ms after each beat mark. We performed baseline correction for each EEG signal band by subtracting the mean value calculated from the 50 ms segments between beat marks. The second row represents brain activity for a segment from the movie \textit{Back to 1942}, recorded from 100 ms before to 180 ms after each beat mark, with baseline correction applied every 20 ms. The third row shows brain activity for a segment from the movie \textit{World Heritage in China}, recorded from 55 ms before to 130 ms after each beat mark, with baseline correction applied every 25 ms. This detailed breakdown demonstrates the dynamic changes in brain activity elicited by different emotional stimuli across multiple time points.

\begin{figure}[h]
	\centering
	\includegraphics[width=0.5\textwidth]{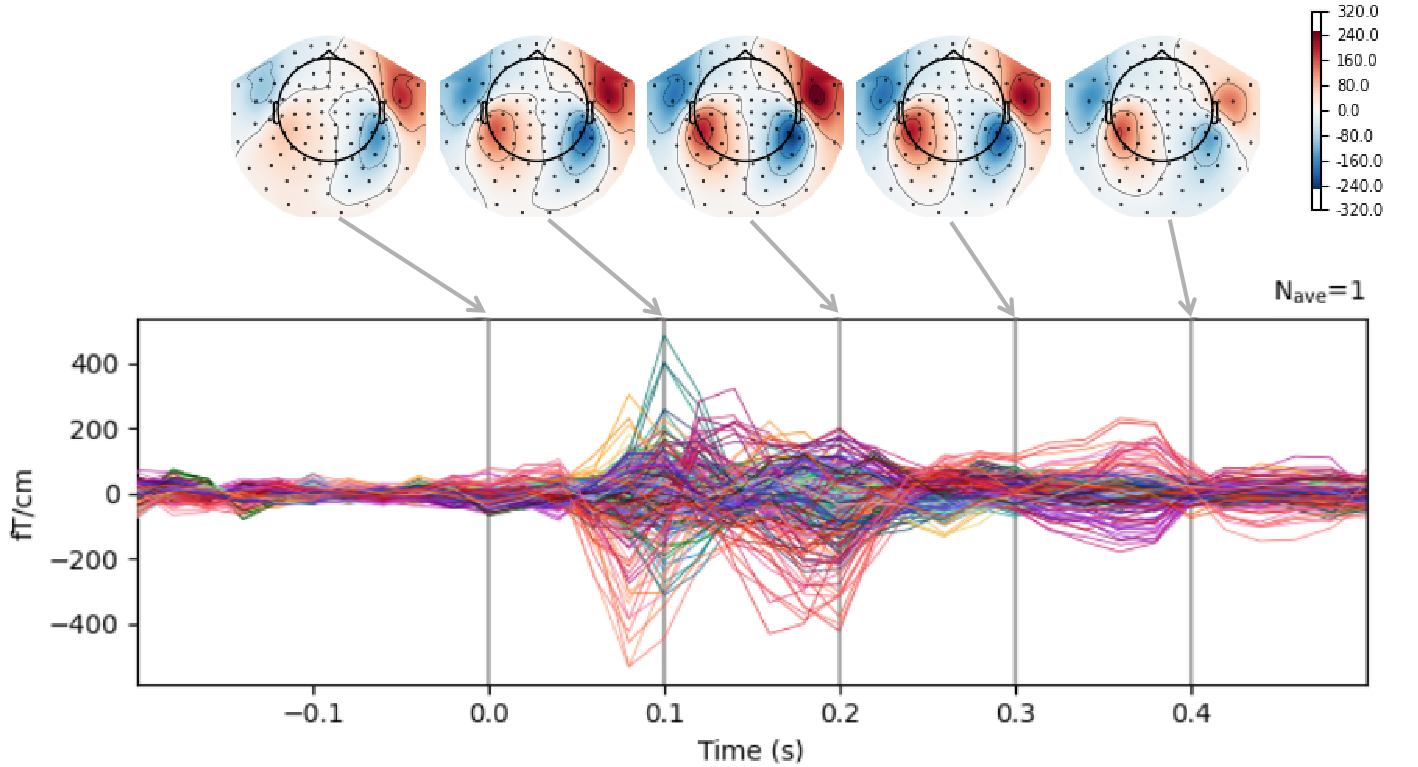}
	\caption{EEG Signal Variations at Time Points in the Movie Scene "Back to 1942" from the SEED Dataset.}
	\label{f7}
\end{figure}

When recording EEG data, the signal initially shows a negative dip at 0 ms. Any auditory processing related to the music beats occurs shortly afterward. One possible explanation is that the sudden change in the music beat causes a sudden dip in the EEG signal. However, this requires further investigation. It could also be due to inaccuracies in the wearable device used to collect the signal or the sampling rate and tracker precision. For further comparison, Figure~\ref{f7}  shows the EEG signal changes at the imagined time points corresponding to different movie scenes in the SEED dataset. This figure corresponds to the second row in Figure~\ref{f6}, showing the variations in the ROIs. The amplitude range of these EEG signal changes is relatively large, likely due to the emotional intensity provided by the movie segments. To calculate meaningful fluctuations, precise timing (e.g., sudden events) must be known. However, small changes during EEG signal recording are likely, causing some imprecision in the beat marks.

To understand the event-related effects on EEG signals induced by auditory and visual stimuli, we used the DEAP dataset, which contains emotion-related EEG signals. We analyzed the five EEG frequency bands affected by emotional states (as shown in Table~\ref{tab1}). We employed Event-Related Potentials (ERP) metrics to record these changes. ERP refers to the specific potentials induced by stimuli such as auditory and visual cues, excluding spontaneous EEG signals. With its high temporal resolution, ERP can reflect changes in brain neurophysiological activities during cognitive processes, making it a widely used tool for assessing brain cognitive processing. The features commonly analyzed in ERP data are the latency and peaks of its components. The main components of ERP include P100, N100, P200, N200, and P300. Most studies focus on the P300 component of ERP data for brainwave analysis. The P300 component is a positive deflection occurring around 300 milliseconds post-stimulus. Table~\ref{tab3} presents the statistical results of P300 values from ERP data in the DEAP dataset. It is evident from the table that participants exhibit more intense emotional reactions to sad music scenes.

\begin{table}[]
\caption{Statistics on the P300 value of ERP data on the DEAP dataset.}
\begin{tabular}{l|c|c}
\hline
\textbf{Emotion Change} & \textbf{\begin{tabular}[c]{@{}c@{}}ERP Change Period \\ Post-Stimulus (ms)\end{tabular}} & \textbf{P} \\ \hline
Happiness (mean$\pm$SD) & 298.45$\pm$33.25                                                                         & 0.026      \\
Sadness (mean$\pm$SD)   & 316.45$\pm$30.24                                                                         & 0.470      \\
Fear (mean$\pm$SD)      & 327.13$\pm$31.69                                                                         & 0.020      \\ \hline
\end{tabular}
\label{tab3}
\end{table}

\begin{figure*}[h]
	\centering
	\includegraphics[width=1.0\textwidth]{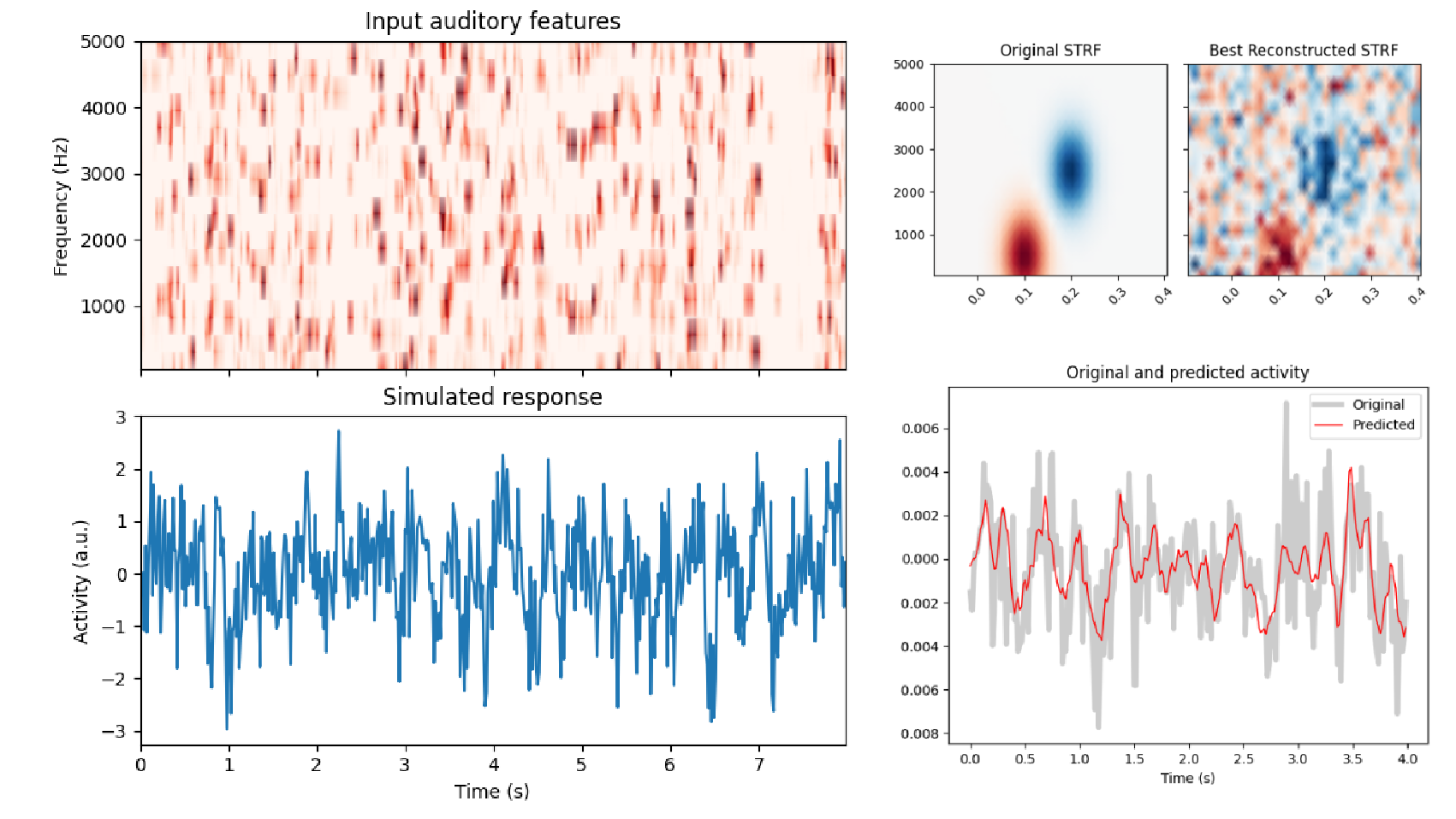}
	\caption{Neural Response Signal Changes and Bi-LSTM Simulated Signal Variation Trends in EEG Data.}
	\label{f8}
\end{figure*}

We used a Bi-LSTM model to simulate and predict EEG signal fluctuations recorded in the DEAP dataset. The Data$\_$preprocessed files were fed into a linear filter to map the spectral output of the EEG signals, representing brain neural activity. We attempted to recreate the receptive field model of the linear filter to generate this data. The data were downsampled along the time dimension, and this receptive field was used to create an artificial neural response. This involved performing a dot product with the receptive field. As shown in Figure~\ref{f8}, the left side depicts the simulated neural response to EEG signals, with auditory features as input and the simulated time-varying stimulus response as output. The top right shows the receptive field generated by the stimulus signals, and the bottom right illustrates the EEG signal variation process simulated by the Bi-LSTM. The figure indicates that the EEG signals exhibit significant fluctuations in response to varying levels of musical stimuli, with the extent of fluctuations being related to the emotional intensity provided by the music.

\section{Conclusion}\label{sec5}

This study records and analyzes brain activity while listening to music using wearable EEG devices and Bi-LSTM with attention mechanisms to identify emotional states. Data was collected using wearable EEG devices, preprocessed, and DE features were extracted. A Bi-LSTM model with attention mechanisms was then constructed, showing superior performance in emotion recognition, particularly with the SEED and DEAP datasets. The Bi-LSTM-AttGW model achieved the highest accuracy of 98.28\% in multi-class tasks.

The novelty of this study lies in the integration of Bi-LSTM with attention mechanisms for EEG signal processing. This combination significantly improves the accuracy of emotion recognition, outperforming traditional models like SVM, EEG-Net, and LSTM~\cite{wang2023computational,luo2023aq2pnn}. The study demonstrates the potential of using advanced deep learning techniques to better capture the complexities of EEG signals and enhance emotion recognition.

Despite these achievements, there are limitations. Wearable EEG devices have relatively low sampling rates and are prone to artifacts, affecting signal quality. Future research should focus on improving device design and signal processing algorithms. Additionally, larger sample sizes and more diverse experimental conditions are needed to account for individual differences in emotional responses to music. Further integration of multimodal data and various emotional elicitation methods could enhance the accuracy and applicability of emotion recognition in BCI and affective computing.

\section*{Author contributions}
\textbf{Jingyi Wang:} Data Management, Methodology, Software, and Writing – original draft. \textbf{Zhiqun Wang:} Supervision, Writing – review \& editing. Fanyu Kong: Methodology, Software, Writing – review \& editing. \textbf{Guiran Liu:} Supervision, Writing – review \& editing.

\section*{Declaration of competing interest}
The authors declare that they have no known competing financial interests or personal relationships that could have appeared to influence the work reported in this paper.

\section*{Data availability}
The data that support the findings of this study are available on request from the corresponding author. The data are not publicly available due to privacy or ethical restrictions.


\bibliographystyle{model1-num-names}
\bibliography{cas-refs}

\begin{thebibliography}{82}
\expandafter\ifx\csname natexlab\endcsname\relax\def\natexlab#1{#1}\fi
\providecommand{\url}[1]{\texttt{#1}}
\providecommand{\href}[2]{#2}
\providecommand{\path}[1]{#1}
\providecommand{\DOIprefix}{doi:}
\providecommand{\ArXivprefix}{arXiv:}
\providecommand{\URLprefix}{URL: }
\providecommand{\Pubmedprefix}{pmid:}
\providecommand{\doi}[1]{\href{http://dx.doi.org/#1}{\path{#1}}}
\providecommand{\Pubmed}[1]{\href{pmid:#1}{\path{#1}}}
\providecommand{\bibinfo}[2]{#2}
\ifx\xfnm\relax \def\xfnm[#1]{\unskip,\space#1}\fi
\bibitem[{Hu et~al.(2020)Hu, Yuan, Xu, Leng, Yuan, and Yuan}]{hu2020scalp}
\bibinfo{author}{X.~Hu}, \bibinfo{author}{S.~Yuan}, \bibinfo{author}{F.~Xu}, \bibinfo{author}{Y.~Leng}, \bibinfo{author}{K.~Yuan}, \bibinfo{author}{Q.~Yuan},
\newblock \bibinfo{title}{Scalp eeg classification using deep bi-lstm network for seizure detection},
\newblock \bibinfo{journal}{Computers in Biology and Medicine} \bibinfo{volume}{124} (\bibinfo{year}{2020}) \bibinfo{pages}{103919}.
\bibitem[{Wang et~al.(2024)Wang, Sun, and Guo}]{SXGY202402006}
\bibinfo{author}{Y.~Wang}, \bibinfo{author}{S.~Sun}, \bibinfo{author}{Q.~Guo},
\newblock \bibinfo{title}{The mechanism of the impact of enterprise digital transformation on transaction performance},
\newblock \bibinfo{journal}{Journal of Xi'an University of Finance and Economics} \bibinfo{volume}{37} (\bibinfo{year}{2024}) \bibinfo{pages}{60--71}.
\bibitem[{Wang et~al.(2022)Wang, Song, Tao, Liotta, Yang, Li, Gao, Sun, Ge, Zhang et~al.}]{wang2022systematic}
\bibinfo{author}{Y.~Wang}, \bibinfo{author}{W.~Song}, \bibinfo{author}{W.~Tao}, \bibinfo{author}{A.~Liotta}, \bibinfo{author}{D.~Yang}, \bibinfo{author}{X.~Li}, \bibinfo{author}{S.~Gao}, \bibinfo{author}{Y.~Sun}, \bibinfo{author}{W.~Ge}, \bibinfo{author}{W.~Zhang}, et~al.,
\newblock \bibinfo{title}{A systematic review on affective computing: Emotion models, databases, and recent advances},
\newblock \bibinfo{journal}{Information Fusion} \bibinfo{volume}{83} (\bibinfo{year}{2022}) \bibinfo{pages}{19--52}.
\bibitem[{Kamble and Sengupta(2021)}]{kamble2021ensemble}
\bibinfo{author}{K.~S. Kamble}, \bibinfo{author}{J.~Sengupta},
\newblock \bibinfo{title}{Ensemble machine learning-based affective computing for emotion recognition using dual-decomposed eeg signals},
\newblock \bibinfo{journal}{IEEE Sensors Journal} \bibinfo{volume}{22} (\bibinfo{year}{2021}) \bibinfo{pages}{2496--2507}.
\bibitem[{Naser and Saha(2021)}]{naser2021influence}
\bibinfo{author}{D.~S. Naser}, \bibinfo{author}{G.~Saha},
\newblock \bibinfo{title}{Influence of music liking on eeg based emotion recognition},
\newblock \bibinfo{journal}{Biomedical Signal Processing and Control} \bibinfo{volume}{64} (\bibinfo{year}{2021}) \bibinfo{pages}{102251}.
\bibitem[{Daly(2023)}]{daly2023neural}
\bibinfo{author}{I.~Daly},
\newblock \bibinfo{title}{Neural decoding of music from the eeg},
\newblock \bibinfo{journal}{Scientific Reports} \bibinfo{volume}{13} (\bibinfo{year}{2023}) \bibinfo{pages}{624}.
\bibitem[{Rahman et~al.(2021)Rahman, Sarkar, Hossain, Hossain, Islam, Hossain, Quinn, and Moni}]{rahman2021recognition}
\bibinfo{author}{M.~M. Rahman}, \bibinfo{author}{A.~K. Sarkar}, \bibinfo{author}{M.~A. Hossain}, \bibinfo{author}{M.~S. Hossain}, \bibinfo{author}{M.~R. Islam}, \bibinfo{author}{M.~B. Hossain}, \bibinfo{author}{J.~M. Quinn}, \bibinfo{author}{M.~A. Moni},
\newblock \bibinfo{title}{Recognition of human emotions using eeg signals: A review},
\newblock \bibinfo{journal}{Computers in biology and medicine} \bibinfo{volume}{136} (\bibinfo{year}{2021}) \bibinfo{pages}{104696}.
\bibitem[{Zheng and Chen(2021)}]{zheng2021attention}
\bibinfo{author}{X.~Zheng}, \bibinfo{author}{W.~Chen},
\newblock \bibinfo{title}{An attention-based bi-lstm method for visual object classification via eeg},
\newblock \bibinfo{journal}{Biomedical Signal Processing and Control} \bibinfo{volume}{63} (\bibinfo{year}{2021}) \bibinfo{pages}{102174}.
\bibitem[{Dai et~al.(2024)Dai, Li, Luo, Zhao, Hong, Zhu, and Liu}]{dai2024ai}
\bibinfo{author}{S.~Dai}, \bibinfo{author}{K.~Li}, \bibinfo{author}{Z.~Luo}, \bibinfo{author}{P.~Zhao}, \bibinfo{author}{B.~Hong}, \bibinfo{author}{A.~Zhu}, \bibinfo{author}{J.~Liu},
\newblock \bibinfo{title}{Ai-based nlp section discusses the application and effect of bag-of-words models and tf-idf in nlp tasks},
\newblock \bibinfo{journal}{Journal of Artificial Intelligence General science (JAIGS) ISSN: 3006-4023} \bibinfo{volume}{5} (\bibinfo{year}{2024}) \bibinfo{pages}{13--21}.
\bibitem[{Wang et~al.(2021)Wang, Alangari, Hihath, Das, and Anantram}]{wang2021machine}
\bibinfo{author}{Y.~Wang}, \bibinfo{author}{M.~Alangari}, \bibinfo{author}{J.~Hihath}, \bibinfo{author}{A.~K. Das}, \bibinfo{author}{M.~Anantram},
\newblock \bibinfo{title}{A machine learning approach for accurate and real-time dna sequence identification},
\newblock \bibinfo{journal}{BMC genomics} \bibinfo{volume}{22} (\bibinfo{year}{2021}) \bibinfo{pages}{1--10}.
\bibitem[{Richardson et~al.(2024)Richardson, Wang, Dubey, and Sprinkle}]{richardson2024reinforcement}
\bibinfo{author}{A.~Richardson}, \bibinfo{author}{X.~Wang}, \bibinfo{author}{A.~Dubey}, \bibinfo{author}{J.~Sprinkle},
\newblock \bibinfo{title}{Reinforcement learning with communication latency with application to stop-and-go wave dissipation},
\newblock in: \bibinfo{booktitle}{2024 IEEE Intelligent Vehicles Symposium (IV)}, \bibinfo{organization}{IEEE}, \bibinfo{year}{2024}, pp. \bibinfo{pages}{1187--1193}.
\bibitem[{Bouallegue et~al.(2020)Bouallegue, Djemal, Alshebeili, and Aldhalaan}]{bouallegue2020dynamic}
\bibinfo{author}{G.~Bouallegue}, \bibinfo{author}{R.~Djemal}, \bibinfo{author}{S.~A. Alshebeili}, \bibinfo{author}{H.~Aldhalaan},
\newblock \bibinfo{title}{A dynamic filtering df-rnn deep-learning-based approach for eeg-based neurological disorders diagnosis},
\newblock \bibinfo{journal}{IEEE Access} \bibinfo{volume}{8} (\bibinfo{year}{2020}) \bibinfo{pages}{206992--207007}.
\bibitem[{Wang et~al.(2024)Wang, Li, An, Zhang, and Sun}]{10438483}
\bibinfo{author}{J.~Wang}, \bibinfo{author}{F.~Li}, \bibinfo{author}{Y.~An}, \bibinfo{author}{X.~Zhang}, \bibinfo{author}{H.~Sun},
\newblock \bibinfo{title}{Towards robust lidar-camera fusion in bev space via mutual deformable attention and temporal aggregation},
\newblock \bibinfo{journal}{IEEE Transactions on Circuits and Systems for Video Technology}  (\bibinfo{year}{2024}) \bibinfo{pages}{1--1}.
\bibitem[{Xu et~al.(2022)Xu, Deng, Dong, and Shimada}]{xu2022dpmpc}
\bibinfo{author}{Z.~Xu}, \bibinfo{author}{D.~Deng}, \bibinfo{author}{Y.~Dong}, \bibinfo{author}{K.~Shimada},
\newblock \bibinfo{title}{Dpmpc-planner: A real-time uav trajectory planning framework for complex static environments with dynamic obstacles},
\newblock in: \bibinfo{booktitle}{2022 International Conference on Robotics and Automation (ICRA)}, \bibinfo{organization}{IEEE}, \bibinfo{year}{2022}, pp. \bibinfo{pages}{250--256}.
\bibitem[{Zhao et~al.(2024)Zhao, Li, Hong, Zhu, Liu, and Dai}]{zhao2024task}
\bibinfo{author}{P.~Zhao}, \bibinfo{author}{K.~Li}, \bibinfo{author}{B.~Hong}, \bibinfo{author}{A.~Zhu}, \bibinfo{author}{J.~Liu}, \bibinfo{author}{S.~Dai},
\newblock \bibinfo{title}{Task allocation planning based on hierarchical task network for national economic mobilization},
\newblock \bibinfo{journal}{Journal of Artificial Intelligence General science (JAIGS) ISSN: 3006-4023} \bibinfo{volume}{5} (\bibinfo{year}{2024}) \bibinfo{pages}{22--31}.
\bibitem[{Song et~al.(2024)Song, Fellegara, Iuricich, and De~Floriani}]{song2024parallel}
\bibinfo{author}{Y.~Song}, \bibinfo{author}{R.~Fellegara}, \bibinfo{author}{F.~Iuricich}, \bibinfo{author}{L.~De~Floriani},
\newblock \bibinfo{title}{Parallel topology-aware mesh simplification on terrain trees},
\newblock \bibinfo{journal}{ACM Transactions on Spatial Algorithms and Systems} \bibinfo{volume}{10} (\bibinfo{year}{2024}) \bibinfo{pages}{1--39}.
\bibitem[{Geng et~al.(2024)Geng, Zhang, Yue, Hu, Wang, Zhang, Yu, Long, and Yan}]{GENG202438}
\bibinfo{author}{X.~Geng}, \bibinfo{author}{X.~Zhang}, \bibinfo{author}{M.~Yue}, \bibinfo{author}{W.~Hu}, \bibinfo{author}{L.~Wang}, \bibinfo{author}{X.~Zhang}, \bibinfo{author}{P.~Yu}, \bibinfo{author}{D.~Long}, \bibinfo{author}{H.~Yan},
\newblock \bibinfo{title}{A motor imagery eeg signal optimized processing algorithm},
\newblock \bibinfo{journal}{Alexandria Engineering Journal} \bibinfo{volume}{101} (\bibinfo{year}{2024}) \bibinfo{pages}{38--51}.
\bibitem[{Zhang et~al.(2024)Zhang, Qi, Zheng, and Shen}]{zhang2024cunet}
\bibinfo{author}{Q.~Zhang}, \bibinfo{author}{W.~Qi}, \bibinfo{author}{H.~Zheng}, \bibinfo{author}{X.~Shen}, \bibinfo{title}{Cu-net: a u-net architecture for efficient brain-tumor segmentation on brats 2019 dataset}, \bibinfo{year}{2024}. \href{http://arxiv.org/abs/2406.13113}{\tt arXiv:2406.13113}.
\bibitem[{Peng et~al.(????)Peng, Ran, Luo, Zhao, Huang, Thorat, Geng, Wang, Xu, Wen et~al.}]{penglingcn}
\bibinfo{author}{H.~Peng}, \bibinfo{author}{R.~Ran}, \bibinfo{author}{Y.~Luo}, \bibinfo{author}{J.~Zhao}, \bibinfo{author}{S.~Huang}, \bibinfo{author}{K.~Thorat}, \bibinfo{author}{T.~Geng}, \bibinfo{author}{C.~Wang}, \bibinfo{author}{X.~Xu}, \bibinfo{author}{W.~Wen}, et~al.,
\newblock \bibinfo{title}{Lingcn: Structural linearized graph convolutional network for homomorphically encrypted inference},
\newblock in: \bibinfo{booktitle}{Thirty-seventh Conference on Neural Information Processing Systems}, ????
\bibitem[{Song et~al.(2021)Song, Fellegara, Iuricich, and De~Floriani}]{song2021efficient}
\bibinfo{author}{Y.~Song}, \bibinfo{author}{R.~Fellegara}, \bibinfo{author}{F.~Iuricich}, \bibinfo{author}{L.~De~Floriani},
\newblock \bibinfo{title}{Efficient topology-aware simplification of large triangulated terrains},
\newblock in: \bibinfo{booktitle}{Proceedings of the 29th International Conference on Advances in Geographic Information Systems}, \bibinfo{year}{2021}, pp. \bibinfo{pages}{576--587}.
\bibitem[{Wang et~al.(2022)Wang, Khandelwal, Das, and Anantram}]{wang2022classification}
\bibinfo{author}{Y.~Wang}, \bibinfo{author}{V.~Khandelwal}, \bibinfo{author}{A.~K. Das}, \bibinfo{author}{M.~Anantram},
\newblock \bibinfo{title}{Classification of dna sequences: Performance evaluation of multiple machine learning methods},
\newblock in: \bibinfo{booktitle}{2022 IEEE 22nd International Conference on Nanotechnology (NANO)}, \bibinfo{organization}{IEEE}, \bibinfo{year}{2022}, pp. \bibinfo{pages}{333--336}.
\bibitem[{Zanetti et~al.(2021)Zanetti, Arza, Aminifar, and Atienza}]{zanetti2021real}
\bibinfo{author}{R.~Zanetti}, \bibinfo{author}{A.~Arza}, \bibinfo{author}{A.~Aminifar}, \bibinfo{author}{D.~Atienza},
\newblock \bibinfo{title}{Real-time eeg-based cognitive workload monitoring on wearable devices},
\newblock \bibinfo{journal}{IEEE transactions on biomedical engineering} \bibinfo{volume}{69} (\bibinfo{year}{2021}) \bibinfo{pages}{265--277}.
\bibitem[{Srinivas et~al.(2024)Srinivas, Arulprakash, Vadivel, Anusha, Rajasekar, and Srinivasan}]{srinivas2024support}
\bibinfo{author}{P.~Srinivas}, \bibinfo{author}{M.~Arulprakash}, \bibinfo{author}{M.~Vadivel}, \bibinfo{author}{N.~Anusha}, \bibinfo{author}{G.~Rajasekar}, \bibinfo{author}{C.~Srinivasan},
\newblock \bibinfo{title}{Support vector machines based predictive seizure care using iot-wearable eeg devices for proactive intervention in epilepsy},
\newblock in: \bibinfo{booktitle}{2024 2nd International Conference on Computer, Communication and Control (IC4)}, \bibinfo{organization}{IEEE}, \bibinfo{year}{2024}, pp. \bibinfo{pages}{1--5}.
\bibitem[{Kotowski et~al.(2018)Kotowski, Stapor, Leski, and Kotas}]{kotowski2018validation}
\bibinfo{author}{K.~Kotowski}, \bibinfo{author}{K.~Stapor}, \bibinfo{author}{J.~Leski}, \bibinfo{author}{M.~Kotas},
\newblock \bibinfo{title}{Validation of emotiv epoc+ for extracting erp correlates of emotional face processing},
\newblock \bibinfo{journal}{Biocybernetics and Biomedical Engineering} \bibinfo{volume}{38} (\bibinfo{year}{2018}) \bibinfo{pages}{773--781}.
\bibitem[{Cheng and Song(2023)}]{SXGY202306005}
\bibinfo{author}{Q.~Cheng}, \bibinfo{author}{Y.~Song},
\newblock \bibinfo{title}{The impact of the digital economy on regional economic development disparities from the perspective of spatial spillovers},
\newblock \bibinfo{journal}{Journal of Xi'an University of Finance and Economics} \bibinfo{volume}{36} (\bibinfo{year}{2023}) \bibinfo{pages}{44--57}.
\bibitem[{Shin et~al.(2022)Shin, Kwon, Kim, Ryu, Ok, Joon~Kwon, Park, and Kim}]{shin2022wearable}
\bibinfo{author}{J.~H. Shin}, \bibinfo{author}{J.~Kwon}, \bibinfo{author}{J.~U. Kim}, \bibinfo{author}{H.~Ryu}, \bibinfo{author}{J.~Ok}, \bibinfo{author}{S.~Joon~Kwon}, \bibinfo{author}{H.~Park}, \bibinfo{author}{T.-i. Kim},
\newblock \bibinfo{title}{Wearable eeg electronics for a brain--ai closed-loop system to enhance autonomous machine decision-making},
\newblock \bibinfo{journal}{npj Flexible Electronics} \bibinfo{volume}{6} (\bibinfo{year}{2022}) \bibinfo{pages}{32}.
\bibitem[{Li et~al.(2022)Li, Zhao, Liu, Liu, and Li}]{li2022effect}
\bibinfo{author}{T.~Li}, \bibinfo{author}{R.~Zhao}, \bibinfo{author}{Y.~Liu}, \bibinfo{author}{X.~Liu}, \bibinfo{author}{Y.~Li},
\newblock \bibinfo{title}{Effect of age on driving behavior and a neurophysiological interpretation},
\newblock in: \bibinfo{booktitle}{International Conference on Human-Computer Interaction}, \bibinfo{organization}{Springer}, \bibinfo{year}{2022}, pp. \bibinfo{pages}{184--194}.
\bibitem[{De et~al.(2022)De, Mohammad, Wang, Kubendran, Das, and Anantram}]{de2022modeling}
\bibinfo{author}{A.~De}, \bibinfo{author}{H.~Mohammad}, \bibinfo{author}{Y.~Wang}, \bibinfo{author}{R.~Kubendran}, \bibinfo{author}{A.~K. Das}, \bibinfo{author}{M.~Anantram},
\newblock \bibinfo{title}{Modeling and simulation of dna origami based electronic read-only memory},
\newblock in: \bibinfo{booktitle}{2022 IEEE 22nd International Conference on Nanotechnology (NANO)}, \bibinfo{organization}{IEEE}, \bibinfo{year}{2022}, pp. \bibinfo{pages}{385--388}.
\bibitem[{Hong et~al.(2024)Hong, Zhao, Liu, Zhu, Dai, and Li}]{hong2024application}
\bibinfo{author}{B.~Hong}, \bibinfo{author}{P.~Zhao}, \bibinfo{author}{J.~Liu}, \bibinfo{author}{A.~Zhu}, \bibinfo{author}{S.~Dai}, \bibinfo{author}{K.~Li},
\newblock \bibinfo{title}{The application of artificial intelligence technology in assembly techniques within the industrial sector},
\newblock \bibinfo{journal}{Journal of Artificial Intelligence General science (JAIGS) ISSN: 3006-4023} \bibinfo{volume}{5} (\bibinfo{year}{2024}) \bibinfo{pages}{1--12}.
\bibitem[{Jin et~al.(2023)Jin, Huang, Zhang, Pechenizkiy, Liu, Liu, and Chen}]{jin2023visual}
\bibinfo{author}{C.~Jin}, \bibinfo{author}{T.~Huang}, \bibinfo{author}{Y.~Zhang}, \bibinfo{author}{M.~Pechenizkiy}, \bibinfo{author}{S.~Liu}, \bibinfo{author}{S.~Liu}, \bibinfo{author}{T.~Chen},
\newblock \bibinfo{title}{Visual prompting upgrades neural network sparsification: A data-model perspective},
\newblock \bibinfo{journal}{arXiv preprint arXiv:2312.01397}  (\bibinfo{year}{2023}).
\bibitem[{Dong(2024)}]{dong2024design}
\bibinfo{author}{Y.~Dong},
\newblock \bibinfo{title}{The design of autonomous uav prototypes for inspecting tunnel construction environment},
\newblock \bibinfo{journal}{arXiv preprint arXiv:2408.07286}  (\bibinfo{year}{2024}).
\bibitem[{Zhao et~al.(2022)Zhao, Liu, Li, and Li}]{zhao2022preliminary}
\bibinfo{author}{R.~Zhao}, \bibinfo{author}{Y.~Liu}, \bibinfo{author}{T.~Li}, \bibinfo{author}{Y.~Li},
\newblock \bibinfo{title}{A preliminary evaluation of driver’s workload in partially automated vehicles},
\newblock in: \bibinfo{booktitle}{International Conference on Human-Computer Interaction}, \bibinfo{organization}{Springer}, \bibinfo{year}{2022}, pp. \bibinfo{pages}{448--458}.
\bibitem[{Jin et~al.(2024)Jin, Che, Peng, Li, and Pavone}]{jin2024learning}
\bibinfo{author}{C.~Jin}, \bibinfo{author}{T.~Che}, \bibinfo{author}{H.~Peng}, \bibinfo{author}{Y.~Li}, \bibinfo{author}{M.~Pavone},
\newblock \bibinfo{title}{Learning from teaching regularization: Generalizable correlations should be easy to imitate},
\newblock \bibinfo{journal}{arXiv preprint arXiv:2402.02769}  (\bibinfo{year}{2024}).
\bibitem[{Jiang et~al.(2020)Jiang, Yu, Qin, Zhuang, Zhang, Hu, and Wu}]{jiang2020dualvd}
\bibinfo{author}{X.~Jiang}, \bibinfo{author}{J.~Yu}, \bibinfo{author}{Z.~Qin}, \bibinfo{author}{Y.~Zhuang}, \bibinfo{author}{X.~Zhang}, \bibinfo{author}{Y.~Hu}, \bibinfo{author}{Q.~Wu},
\newblock \bibinfo{title}{Dualvd: An adaptive dual encoding model for deep visual understanding in visual dialogue},
\newblock in: \bibinfo{booktitle}{Proceedings of the AAAI conference on artificial intelligence}, volume~\bibinfo{volume}{34}, \bibinfo{year}{2020}, pp. \bibinfo{pages}{11125--11132}.
\bibitem[{Li et~al.(2021)Li, Zhao, Liu, Li, and Li}]{li2021evaluate}
\bibinfo{author}{T.~Li}, \bibinfo{author}{R.~Zhao}, \bibinfo{author}{Y.~Liu}, \bibinfo{author}{Y.~Li}, \bibinfo{author}{G.~Li},
\newblock \bibinfo{title}{Evaluate the effect of age and driving experience on driving performance with automated vehicles},
\newblock in: \bibinfo{booktitle}{International Conference on Applied Human Factors and Ergonomics}, \bibinfo{organization}{Springer}, \bibinfo{year}{2021}, pp. \bibinfo{pages}{155--161}.
\bibitem[{Xie et~al.(2023)Xie, Peng, Hasan, Huang, Zhao, Fang, Zhang, Geng, Khan, and Ding}]{xie2023accel}
\bibinfo{author}{X.~Xie}, \bibinfo{author}{H.~Peng}, \bibinfo{author}{A.~Hasan}, \bibinfo{author}{S.~Huang}, \bibinfo{author}{J.~Zhao}, \bibinfo{author}{H.~Fang}, \bibinfo{author}{W.~Zhang}, \bibinfo{author}{T.~Geng}, \bibinfo{author}{O.~Khan}, \bibinfo{author}{C.~Ding},
\newblock \bibinfo{title}{Accel-gcn: High-performance gpu accelerator design for graph convolution networks},
\newblock in: \bibinfo{booktitle}{2023 IEEE/ACM International Conference on Computer Aided Design (ICCAD)}, \bibinfo{organization}{IEEE}, \bibinfo{year}{2023}, pp. \bibinfo{pages}{01--09}.
\bibitem[{Chen et~al.(2024)Chen, Zhang, Dong, Zhou, and Wang}]{chen2024enhancing}
\bibinfo{author}{P.~Chen}, \bibinfo{author}{Z.~Zhang}, \bibinfo{author}{Y.~Dong}, \bibinfo{author}{L.~Zhou}, \bibinfo{author}{H.~Wang},
\newblock \bibinfo{title}{Enhancing visual question answering through ranking-based hybrid training and multimodal fusion},
\newblock \bibinfo{journal}{arXiv preprint arXiv:2408.07303}  (\bibinfo{year}{2024}).
\bibitem[{Li et~al.(2024)Li, Zhu, Zhou, Zhao, Song, and Liu}]{li2024utilizing}
\bibinfo{author}{K.~Li}, \bibinfo{author}{A.~Zhu}, \bibinfo{author}{W.~Zhou}, \bibinfo{author}{P.~Zhao}, \bibinfo{author}{J.~Song}, \bibinfo{author}{J.~Liu},
\newblock \bibinfo{title}{Utilizing deep learning to optimize software development processes},
\newblock \bibinfo{journal}{arXiv preprint arXiv:2404.13630}  (\bibinfo{year}{2024}).
\bibitem[{Lambrecq et~al.(2021)Lambrecq, Hanin, Munoz-Musat, Chougar, Gassama, Delorme, Cousyn, Borden, Damiano, Frazzini et~al.}]{lambrecq2021association}
\bibinfo{author}{V.~Lambrecq}, \bibinfo{author}{A.~Hanin}, \bibinfo{author}{E.~Munoz-Musat}, \bibinfo{author}{L.~Chougar}, \bibinfo{author}{S.~Gassama}, \bibinfo{author}{C.~Delorme}, \bibinfo{author}{L.~Cousyn}, \bibinfo{author}{A.~Borden}, \bibinfo{author}{M.~Damiano}, \bibinfo{author}{V.~Frazzini}, et~al.,
\newblock \bibinfo{title}{Association of clinical, biological, and brain magnetic resonance imaging findings with electroencephalographic findings for patients with covid-19},
\newblock \bibinfo{journal}{JAMA Network Open} \bibinfo{volume}{4} (\bibinfo{year}{2021}) \bibinfo{pages}{e211489--e211489}.
\bibitem[{Nentwich et~al.(2020)Nentwich, Ai, Madsen, Telesford, Haufe, Milham, and Parra}]{nentwich2020functional}
\bibinfo{author}{M.~Nentwich}, \bibinfo{author}{L.~Ai}, \bibinfo{author}{J.~Madsen}, \bibinfo{author}{Q.~K. Telesford}, \bibinfo{author}{S.~Haufe}, \bibinfo{author}{M.~P. Milham}, \bibinfo{author}{L.~C. Parra},
\newblock \bibinfo{title}{Functional connectivity of eeg is subject-specific, associated with phenotype, and different from fmri},
\newblock \bibinfo{journal}{NeuroImage} \bibinfo{volume}{218} (\bibinfo{year}{2020}) \bibinfo{pages}{117001}.
\bibitem[{Palva and Palva(2012)}]{palva2012infra}
\bibinfo{author}{J.~M. Palva}, \bibinfo{author}{S.~Palva},
\newblock \bibinfo{title}{Infra-slow fluctuations in electrophysiological recordings, blood-oxygenation-level-dependent signals, and psychophysical time series},
\newblock \bibinfo{journal}{Neuroimage} \bibinfo{volume}{62} (\bibinfo{year}{2012}) \bibinfo{pages}{2201--2211}.
\bibitem[{Wilde et~al.(2020)Wilde, Goodrich-Hunsaker, Ware, Taylor, Biekman, Hunter, Newman-Norlund, Scarneo, Casa, and Levin}]{wilde2020diffusion}
\bibinfo{author}{E.~A. Wilde}, \bibinfo{author}{N.~J. Goodrich-Hunsaker}, \bibinfo{author}{A.~L. Ware}, \bibinfo{author}{B.~A. Taylor}, \bibinfo{author}{B.~D. Biekman}, \bibinfo{author}{J.~V. Hunter}, \bibinfo{author}{R.~Newman-Norlund}, \bibinfo{author}{S.~Scarneo}, \bibinfo{author}{D.~J. Casa}, \bibinfo{author}{H.~S. Levin},
\newblock \bibinfo{title}{Diffusion tensor imaging indicators of white matter injury are correlated with a multimodal electroencephalography-based biomarker in slow recovering, concussed collegiate athletes},
\newblock \bibinfo{journal}{Journal of neurotrauma} \bibinfo{volume}{37} (\bibinfo{year}{2020}) \bibinfo{pages}{2093--2101}.
\bibitem[{Tortora et~al.(2020)Tortora, Ghidoni, Chisari, Micera, and Artoni}]{tortora2020deep}
\bibinfo{author}{S.~Tortora}, \bibinfo{author}{S.~Ghidoni}, \bibinfo{author}{C.~Chisari}, \bibinfo{author}{S.~Micera}, \bibinfo{author}{F.~Artoni},
\newblock \bibinfo{title}{Deep learning-based bci for gait decoding from eeg with lstm recurrent neural network},
\newblock \bibinfo{journal}{Journal of neural engineering} \bibinfo{volume}{17} (\bibinfo{year}{2020}) \bibinfo{pages}{046011}.
\bibitem[{Teipel et~al.(2009)Teipel, Pogarell, Meindl, Dietrich, Sydykova, Hunklinger, Georgii, Mulert, Reiser, M{\"o}ller et~al.}]{teipel2009regional}
\bibinfo{author}{S.~J. Teipel}, \bibinfo{author}{O.~Pogarell}, \bibinfo{author}{T.~Meindl}, \bibinfo{author}{O.~Dietrich}, \bibinfo{author}{D.~Sydykova}, \bibinfo{author}{U.~Hunklinger}, \bibinfo{author}{B.~Georgii}, \bibinfo{author}{C.~Mulert}, \bibinfo{author}{M.~F. Reiser}, \bibinfo{author}{H.-J. M{\"o}ller}, et~al.,
\newblock \bibinfo{title}{Regional networks underlying interhemispheric connectivity: an eeg and dti study in healthy ageing and amnestic mild cognitive impairment},
\newblock \bibinfo{journal}{Human brain mapping} \bibinfo{volume}{30} (\bibinfo{year}{2009}) \bibinfo{pages}{2098--2119}.
\bibitem[{Cui et~al.(2023)Cui, Liu, Zhang, Chen, Liu, and Chen}]{9786857}
\bibinfo{author}{H.~Cui}, \bibinfo{author}{A.~Liu}, \bibinfo{author}{X.~Zhang}, \bibinfo{author}{X.~Chen}, \bibinfo{author}{J.~Liu}, \bibinfo{author}{X.~Chen},
\newblock \bibinfo{title}{Eeg-based subject-independent emotion recognition using gated recurrent unit and minimum class confusion},
\newblock \bibinfo{journal}{IEEE Transactions on Affective Computing} \bibinfo{volume}{14} (\bibinfo{year}{2023}) \bibinfo{pages}{2740--2750}.
\bibitem[{Li et~al.(2024)Li, Pang, Bai, Zheng, Zhou, and Ning}]{li2024learning}
\bibinfo{author}{T.~Li}, \bibinfo{author}{G.~Pang}, \bibinfo{author}{X.~Bai}, \bibinfo{author}{J.~Zheng}, \bibinfo{author}{L.~Zhou}, \bibinfo{author}{X.~Ning},
\newblock \bibinfo{title}{Learning adversarial semantic embeddings for zero-shot recognition in open worlds},
\newblock \bibinfo{journal}{Pattern Recognition} \bibinfo{volume}{149} (\bibinfo{year}{2024}) \bibinfo{pages}{110258}.
\bibitem[{Ran et~al.(2024)Ran, Li, Li, Tian, Ning, and Tiwari}]{RAN2024103664}
\bibinfo{author}{H.~Ran}, \bibinfo{author}{W.~Li}, \bibinfo{author}{L.~Li}, \bibinfo{author}{S.~Tian}, \bibinfo{author}{X.~Ning}, \bibinfo{author}{P.~Tiwari},
\newblock \bibinfo{title}{Learning optimal inter-class margin adaptively for few-shot class-incremental learning via neural collapse-based meta-learning},
\newblock \bibinfo{journal}{Information Processing \& Management} \bibinfo{volume}{61} (\bibinfo{year}{2024}) \bibinfo{pages}{103664}.
\bibitem[{{Ein Shoka} et~al.(2023){Ein Shoka}, Dessouky, El-Sayed, and {El-Din Hemdan}}]{EINSHOKA2023399}
\bibinfo{author}{A.~A. {Ein Shoka}}, \bibinfo{author}{M.~M. Dessouky}, \bibinfo{author}{A.~El-Sayed}, \bibinfo{author}{E.~{El-Din Hemdan}},
\newblock \bibinfo{title}{An efficient cnn based epileptic seizures detection framework using encrypted eeg signals for secure telemedicine applications},
\newblock \bibinfo{journal}{Alexandria Engineering Journal} \bibinfo{volume}{65} (\bibinfo{year}{2023}) \bibinfo{pages}{399--412}.
\bibitem[{Zhu et~al.(2024)Zhu, Li, Wu, Zhao, Zhou, and Hong}]{zhu2024cross}
\bibinfo{author}{A.~Zhu}, \bibinfo{author}{K.~Li}, \bibinfo{author}{T.~Wu}, \bibinfo{author}{P.~Zhao}, \bibinfo{author}{W.~Zhou}, \bibinfo{author}{B.~Hong},
\newblock \bibinfo{title}{Cross-task multi-branch vision transformer for facial expression and mask wearing classification},
\newblock \bibinfo{journal}{arXiv preprint arXiv:2404.14606}  (\bibinfo{year}{2024}).
\bibitem[{Li et~al.(2024)Li, Xirui, Song, Hong, and Wang}]{li2024application}
\bibinfo{author}{K.~Li}, \bibinfo{author}{P.~Xirui}, \bibinfo{author}{J.~Song}, \bibinfo{author}{B.~Hong}, \bibinfo{author}{J.~Wang},
\newblock \bibinfo{title}{The application of augmented reality (ar) in remote work and education},
\newblock \bibinfo{journal}{arXiv preprint arXiv:2404.10579}  (\bibinfo{year}{2024}).
\bibitem[{An et~al.(2023)An, Wang, T.~Johnson, Sprinkle, and Ma}]{an2023runtime}
\bibinfo{author}{Z.~An}, \bibinfo{author}{X.~Wang}, \bibinfo{author}{T.~T.~Johnson}, \bibinfo{author}{J.~Sprinkle}, \bibinfo{author}{M.~Ma},
\newblock \bibinfo{title}{Runtime monitoring of accidents in driving recordings with multi-type logic in empirical models},
\newblock in: \bibinfo{booktitle}{International Conference on Runtime Verification}, \bibinfo{organization}{Springer}, \bibinfo{year}{2023}, pp. \bibinfo{pages}{376--388}.
\bibitem[{Liu et~al.(2021)Liu, Zhao, Li, and Li}]{liu2021investigation}
\bibinfo{author}{Y.~Liu}, \bibinfo{author}{R.~Zhao}, \bibinfo{author}{T.~Li}, \bibinfo{author}{Y.~Li},
\newblock \bibinfo{title}{An investigation of the impact of autonomous driving on driving behavior in traffic jam},
\newblock in: \bibinfo{booktitle}{IIE Annual Conference. Proceedings}, \bibinfo{organization}{Institute of Industrial and Systems Engineers (IISE)}, \bibinfo{year}{2021}, pp. \bibinfo{pages}{986--991}.
\bibitem[{Sheykhivand et~al.(2020)Sheykhivand, Mousavi, Rezaii, and Farzamnia}]{sheykhivand2020recognizing}
\bibinfo{author}{S.~Sheykhivand}, \bibinfo{author}{Z.~Mousavi}, \bibinfo{author}{T.~Y. Rezaii}, \bibinfo{author}{A.~Farzamnia},
\newblock \bibinfo{title}{Recognizing emotions evoked by music using cnn-lstm networks on eeg signals},
\newblock \bibinfo{journal}{IEEE access} \bibinfo{volume}{8} (\bibinfo{year}{2020}) \bibinfo{pages}{139332--139345}.
\bibitem[{Wang et~al.(2024)Wang, Onwumelu, and Sprinkle}]{wang2024using}
\bibinfo{author}{X.~Wang}, \bibinfo{author}{S.~Onwumelu}, \bibinfo{author}{J.~Sprinkle},
\newblock \bibinfo{title}{Using automated vehicle data as a fitness tracker for sustainability},
\newblock in: \bibinfo{booktitle}{2024 Forum for Innovative Sustainable Transportation Systems (FISTS)}, \bibinfo{organization}{IEEE}, \bibinfo{year}{2024}, pp. \bibinfo{pages}{1--6}.
\bibitem[{Fellegara et~al.(2023)Fellegara, Iuricich, Song, and Floriani}]{fellegara2023terrain}
\bibinfo{author}{R.~Fellegara}, \bibinfo{author}{F.~Iuricich}, \bibinfo{author}{Y.~Song}, \bibinfo{author}{L.~D. Floriani},
\newblock \bibinfo{title}{Terrain trees: a framework for representing, analyzing and visualizing triangulated terrains},
\newblock \bibinfo{journal}{GeoInformatica} \bibinfo{volume}{27} (\bibinfo{year}{2023}) \bibinfo{pages}{525--564}.
\bibitem[{Peng et~al.(2023)Peng, Huang, Zhou, Luo, Wang, Wang, Zhao, Xie, Li, Geng et~al.}]{peng2023autorep}
\bibinfo{author}{H.~Peng}, \bibinfo{author}{S.~Huang}, \bibinfo{author}{T.~Zhou}, \bibinfo{author}{Y.~Luo}, \bibinfo{author}{C.~Wang}, \bibinfo{author}{Z.~Wang}, \bibinfo{author}{J.~Zhao}, \bibinfo{author}{X.~Xie}, \bibinfo{author}{A.~Li}, \bibinfo{author}{T.~Geng}, et~al.,
\newblock \bibinfo{title}{Autorep: Automatic relu replacement for fast private network inference},
\newblock in: \bibinfo{booktitle}{2023 IEEE/CVF International Conference on Computer Vision (ICCV)}, \bibinfo{organization}{IEEE}, \bibinfo{year}{2023}, pp. \bibinfo{pages}{5155--5165}.
\bibitem[{Zhuang et~al.(2020)Zhuang, Chen, and Zheng}]{zhuang2020music}
\bibinfo{author}{Y.~Zhuang}, \bibinfo{author}{Y.~Chen}, \bibinfo{author}{J.~Zheng},
\newblock \bibinfo{title}{Music genre classification with transformer classifier},
\newblock in: \bibinfo{booktitle}{Proceedings of the 2020 4th international conference on digital signal processing}, \bibinfo{year}{2020}, pp. \bibinfo{pages}{155--159}.
\bibitem[{Zhao et~al.(2021)Zhao, Liu, Li, and Tokgoz}]{zhao2021investigation}
\bibinfo{author}{R.~Zhao}, \bibinfo{author}{Y.~Liu}, \bibinfo{author}{Y.~Li}, \bibinfo{author}{B.~Tokgoz},
\newblock \bibinfo{title}{An investigation of resilience in manual driving and automatic driving in freight transportation system},
\newblock in: \bibinfo{booktitle}{IIE Annual Conference. Proceedings}, \bibinfo{organization}{Institute of Industrial and Systems Engineers (IISE)}, \bibinfo{year}{2021}, pp. \bibinfo{pages}{974--979}.
\bibitem[{De et~al.(2023)De, Mohammad, Wang, Kubendran, Das, and Anantram}]{de2023performance}
\bibinfo{author}{A.~De}, \bibinfo{author}{H.~Mohammad}, \bibinfo{author}{Y.~Wang}, \bibinfo{author}{R.~Kubendran}, \bibinfo{author}{A.~K. Das}, \bibinfo{author}{M.~Anantram},
\newblock \bibinfo{title}{Performance analysis of dna crossbar arrays for high-density memory storage applications},
\newblock \bibinfo{journal}{Scientific Reports} \bibinfo{volume}{13} (\bibinfo{year}{2023}) \bibinfo{pages}{6650}.
\bibitem[{Liu et~al.(2022)Liu, Zhao, and Li}]{liu2022preliminary}
\bibinfo{author}{Y.~Liu}, \bibinfo{author}{R.~Zhao}, \bibinfo{author}{Y.~Li},
\newblock \bibinfo{title}{A preliminary comparison of drivers’ overtaking behavior between partially automated vehicles and conventional vehicles},
\newblock in: \bibinfo{booktitle}{Proceedings of the Human Factors and Ergonomics Society Annual Meeting}, volume~\bibinfo{volume}{66}, \bibinfo{organization}{SAGE Publications Sage CA: Los Angeles, CA}, \bibinfo{year}{2022}, pp. \bibinfo{pages}{913--917}.
\bibitem[{Somers et~al.(2021)Somers, Long, and Francart}]{somers2021eeg}
\bibinfo{author}{B.~Somers}, \bibinfo{author}{C.~J. Long}, \bibinfo{author}{T.~Francart},
\newblock \bibinfo{title}{Eeg-based diagnostics of the auditory system using cochlear implant electrodes as sensors},
\newblock \bibinfo{journal}{Scientific Reports} \bibinfo{volume}{11} (\bibinfo{year}{2021}) \bibinfo{pages}{5383}.
\bibitem[{Henao et~al.(2020)Henao, Navarrete, Valderrama, and Le~Van~Quyen}]{henao2020entrainment}
\bibinfo{author}{D.~Henao}, \bibinfo{author}{M.~Navarrete}, \bibinfo{author}{M.~Valderrama}, \bibinfo{author}{M.~Le~Van~Quyen},
\newblock \bibinfo{title}{Entrainment and synchronization of brain oscillations to auditory stimulations},
\newblock \bibinfo{journal}{Neuroscience Research} \bibinfo{volume}{156} (\bibinfo{year}{2020}) \bibinfo{pages}{271--278}.
\bibitem[{Ferster et~al.(2022)Ferster, Da~Poian, Menachery, Schreiner, Lustenberger, Maric, Huber, Baumann, and Karlen}]{ferster2022benchmarking}
\bibinfo{author}{M.~L. Ferster}, \bibinfo{author}{G.~Da~Poian}, \bibinfo{author}{K.~Menachery}, \bibinfo{author}{S.~J. Schreiner}, \bibinfo{author}{C.~Lustenberger}, \bibinfo{author}{A.~Maric}, \bibinfo{author}{R.~Huber}, \bibinfo{author}{C.~R. Baumann}, \bibinfo{author}{W.~Karlen},
\newblock \bibinfo{title}{Benchmarking real-time algorithms for in-phase auditory stimulation of low amplitude slow waves with wearable eeg devices during sleep},
\newblock \bibinfo{journal}{IEEE Transactions on Biomedical Engineering} \bibinfo{volume}{69} (\bibinfo{year}{2022}) \bibinfo{pages}{2916--2925}.
\bibitem[{Geng et~al.(2022)Geng, Li, Chen, Yu, Yan, and Yue}]{GENG20224807}
\bibinfo{author}{X.~Geng}, \bibinfo{author}{D.~Li}, \bibinfo{author}{H.~Chen}, \bibinfo{author}{P.~Yu}, \bibinfo{author}{H.~Yan}, \bibinfo{author}{M.~Yue},
\newblock \bibinfo{title}{An improved feature extraction algorithms of eeg signals based on motor imagery brain-computer interface},
\newblock \bibinfo{journal}{Alexandria Engineering Journal} \bibinfo{volume}{61} (\bibinfo{year}{2022}) \bibinfo{pages}{4807--4820}.
\bibitem[{Deng et~al.(2024)Deng, Fan, Li, Pan, Kang, and Zhou}]{deng2024solving}
\bibinfo{author}{Q.~Deng}, \bibinfo{author}{Z.~Fan}, \bibinfo{author}{Z.~Li}, \bibinfo{author}{X.~Pan}, \bibinfo{author}{Q.~Kang}, \bibinfo{author}{M.~Zhou},
\newblock \bibinfo{title}{Solving the food-energy-water nexus problem via intelligent optimization algorithms},
\newblock \bibinfo{journal}{arXiv preprint arXiv:2404.06769}  (\bibinfo{year}{2024}).
\bibitem[{Patel et~al.(2022)Patel, Liu, Zhao, Liu, and Li}]{patel2022inspection}
\bibinfo{author}{S.~Patel}, \bibinfo{author}{Y.~Liu}, \bibinfo{author}{R.~Zhao}, \bibinfo{author}{X.~Liu}, \bibinfo{author}{Y.~Li},
\newblock \bibinfo{title}{Inspection of in-vehicle touchscreen infotainment display for different screen locations, menu types, and positions},
\newblock in: \bibinfo{booktitle}{International conference on human-computer interaction}, \bibinfo{organization}{Springer}, \bibinfo{year}{2022}, pp. \bibinfo{pages}{258--279}.
\bibitem[{Lee et~al.(2024)Lee, Wang, Jang, Hayat, Bunting, Alanqary, Barbour, Fu, Gong, Gunter et~al.}]{lee2024traffic}
\bibinfo{author}{J.~Lee}, \bibinfo{author}{H.~Wang}, \bibinfo{author}{K.~Jang}, \bibinfo{author}{A.~Hayat}, \bibinfo{author}{M.~Bunting}, \bibinfo{author}{A.~Alanqary}, \bibinfo{author}{W.~Barbour}, \bibinfo{author}{Z.~Fu}, \bibinfo{author}{X.~Gong}, \bibinfo{author}{G.~Gunter}, et~al.,
\newblock \bibinfo{title}{Traffic smoothing via connected \& automated vehicles: A modular, hierarchical control design deployed in a 100-cav flow smoothing experiment},
\newblock \bibinfo{journal}{IEEE Control Systems Magazine}  (\bibinfo{year}{2024}).
\bibitem[{Zhu et~al.(2019)Zhu, Zhao, Ni, and Zhang}]{zhu2019image}
\bibinfo{author}{Z.~Zhu}, \bibinfo{author}{R.~Zhao}, \bibinfo{author}{J.~Ni}, \bibinfo{author}{J.~Zhang},
\newblock \bibinfo{title}{Image and spectrum based deep feature analysis for particle matter estimation with weather informatio},
\newblock in: \bibinfo{booktitle}{2019 IEEE International Conference on Image Processing (ICIP)}, \bibinfo{organization}{IEEE}, \bibinfo{year}{2019}, pp. \bibinfo{pages}{3427--3431}.
\bibitem[{Liu et~al.(2022)Liu, Zhao, Li, and Li}]{liu2022impact}
\bibinfo{author}{Y.~Liu}, \bibinfo{author}{R.~Zhao}, \bibinfo{author}{T.~Li}, \bibinfo{author}{Y.~Li},
\newblock \bibinfo{title}{The impact of directional road signs combinations and language unfamiliarity on driving behavior},
\newblock in: \bibinfo{booktitle}{International Conference on Human-Computer Interaction}, \bibinfo{organization}{Springer}, \bibinfo{year}{2022}, pp. \bibinfo{pages}{195--204}.
\bibitem[{Zhang et~al.(2020)Zhang, Wei, Zou, and Fu}]{zhang2020automatic}
\bibinfo{author}{J.~Zhang}, \bibinfo{author}{Z.~Wei}, \bibinfo{author}{J.~Zou}, \bibinfo{author}{H.~Fu},
\newblock \bibinfo{title}{Automatic epileptic eeg classification based on differential entropy and attention model},
\newblock \bibinfo{journal}{Engineering Applications of Artificial Intelligence} \bibinfo{volume}{96} (\bibinfo{year}{2020}) \bibinfo{pages}{103975}.
\bibitem[{Zhou et~al.(2024)Zhou, Zhao, Luo, Xie, Wen, Ding, and Xu}]{zhou2024adapi}
\bibinfo{author}{T.~Zhou}, \bibinfo{author}{J.~Zhao}, \bibinfo{author}{Y.~Luo}, \bibinfo{author}{X.~Xie}, \bibinfo{author}{W.~Wen}, \bibinfo{author}{C.~Ding}, \bibinfo{author}{X.~Xu},
\newblock \bibinfo{title}{Adapi: Facilitating dnn model adaptivity for efficient private inference in edge computing},
\newblock \bibinfo{journal}{arXiv preprint arXiv:2407.05633}  (\bibinfo{year}{2024}).
\bibitem[{Peng et~al.(2024)Peng, Xie, Shivdikar, Hasan, Zhao, Huang, Khan, Kaeli, and Ding}]{peng2024maxk}
\bibinfo{author}{H.~Peng}, \bibinfo{author}{X.~Xie}, \bibinfo{author}{K.~Shivdikar}, \bibinfo{author}{M.~A. Hasan}, \bibinfo{author}{J.~Zhao}, \bibinfo{author}{S.~Huang}, \bibinfo{author}{O.~Khan}, \bibinfo{author}{D.~Kaeli}, \bibinfo{author}{C.~Ding},
\newblock \bibinfo{title}{Maxk-gnn: Extremely fast gpu kernel design for accelerating graph neural networks training},
\newblock in: \bibinfo{booktitle}{Proceedings of the 29th ACM International Conference on Architectural Support for Programming Languages and Operating Systems, Volume 2}, ASPLOS '24, \bibinfo{publisher}{Association for Computing Machinery}, \bibinfo{address}{New York, NY, USA}, \bibinfo{year}{2024}, p. \bibinfo{pages}{683–698}.
\bibitem[{Jin et~al.(2024)Jin, Peng, Zhao, Wang, Xu, Han, Zhao, Zhong, Rajasekaran, and Metaxas}]{jin2024apeer}
\bibinfo{author}{C.~Jin}, \bibinfo{author}{H.~Peng}, \bibinfo{author}{S.~Zhao}, \bibinfo{author}{Z.~Wang}, \bibinfo{author}{W.~Xu}, \bibinfo{author}{L.~Han}, \bibinfo{author}{J.~Zhao}, \bibinfo{author}{K.~Zhong}, \bibinfo{author}{S.~Rajasekaran}, \bibinfo{author}{D.~N. Metaxas},
\newblock \bibinfo{title}{Apeer: Automatic prompt engineering enhances large language model reranking},
\newblock \bibinfo{journal}{arXiv preprint arXiv:2406.14449}  (\bibinfo{year}{2024}).
\bibitem[{Cimtay and Ekmekcioglu(2020)}]{cimtay2020investigating}
\bibinfo{author}{Y.~Cimtay}, \bibinfo{author}{E.~Ekmekcioglu},
\newblock \bibinfo{title}{Investigating the use of pretrained convolutional neural network on cross-subject and cross-dataset eeg emotion recognition},
\newblock \bibinfo{journal}{Sensors} \bibinfo{volume}{20} (\bibinfo{year}{2020}) \bibinfo{pages}{2034}.
\bibitem[{Khateeb et~al.(2021)Khateeb, Anwar, and Alnowami}]{khateeb2021multi}
\bibinfo{author}{M.~Khateeb}, \bibinfo{author}{S.~M. Anwar}, \bibinfo{author}{M.~Alnowami},
\newblock \bibinfo{title}{Multi-domain feature fusion for emotion classification using deap dataset},
\newblock \bibinfo{journal}{Ieee Access} \bibinfo{volume}{9} (\bibinfo{year}{2021}) \bibinfo{pages}{12134--12142}.
\bibitem[{Antony et~al.(2022)Antony, Sankaralingam, Mahendran, Gardezi, Shafiq, Choi, and Hamam}]{antony2022classification}
\bibinfo{author}{M.~J. Antony}, \bibinfo{author}{B.~P. Sankaralingam}, \bibinfo{author}{R.~K. Mahendran}, \bibinfo{author}{A.~A. Gardezi}, \bibinfo{author}{M.~Shafiq}, \bibinfo{author}{J.-G. Choi}, \bibinfo{author}{H.~Hamam},
\newblock \bibinfo{title}{Classification of eeg using adaptive svm classifier with csp and online recursive independent component analysis},
\newblock \bibinfo{journal}{Sensors} \bibinfo{volume}{22} (\bibinfo{year}{2022}) \bibinfo{pages}{7596}.
\bibitem[{Kang et~al.(2020)Kang, Han, Song, Niu, and Li}]{kang2020identification}
\bibinfo{author}{J.~Kang}, \bibinfo{author}{X.~Han}, \bibinfo{author}{J.~Song}, \bibinfo{author}{Z.~Niu}, \bibinfo{author}{X.~Li},
\newblock \bibinfo{title}{The identification of children with autism spectrum disorder by svm approach on eeg and eye-tracking data},
\newblock \bibinfo{journal}{Computers in biology and medicine} \bibinfo{volume}{120} (\bibinfo{year}{2020}) \bibinfo{pages}{103722}.
\bibitem[{Hou et~al.(2023)Hou, Liu, Bai, Yang, Liu, Gao, and Song}]{hou2023eeg}
\bibinfo{author}{F.~Hou}, \bibinfo{author}{J.~Liu}, \bibinfo{author}{Z.~Bai}, \bibinfo{author}{Z.~Yang}, \bibinfo{author}{J.~Liu}, \bibinfo{author}{Q.~Gao}, \bibinfo{author}{Y.~Song},
\newblock \bibinfo{title}{Eeg-based emotion recognition for hearing impaired and normal individuals with residual feature pyramids network based on time--frequency--spatial features},
\newblock \bibinfo{journal}{IEEE Transactions on Instrumentation and Measurement} \bibinfo{volume}{72} (\bibinfo{year}{2023}) \bibinfo{pages}{1--11}.
\bibitem[{Huang et~al.(2022)Huang, Wang, Liu, Li, and Tang}]{huang2022virtual}
\bibinfo{author}{D.~Huang}, \bibinfo{author}{X.~Wang}, \bibinfo{author}{J.~Liu}, \bibinfo{author}{J.~Li}, \bibinfo{author}{W.~Tang},
\newblock \bibinfo{title}{Virtual reality safety training using deep eeg-net and physiology data},
\newblock \bibinfo{journal}{The visual computer} \bibinfo{volume}{38} (\bibinfo{year}{2022}) \bibinfo{pages}{1195--1207}.
\bibitem[{Kadri et~al.(2023)Kadri, Ellouze, Ksantini, and Turki}]{kadri2023new}
\bibinfo{author}{N.~Kadri}, \bibinfo{author}{A.~Ellouze}, \bibinfo{author}{M.~Ksantini}, \bibinfo{author}{S.~H. Turki},
\newblock \bibinfo{title}{New lstm deep learning algorithm for driving behavior classification},
\newblock \bibinfo{journal}{Cybernetics and Systems} \bibinfo{volume}{54} (\bibinfo{year}{2023}) \bibinfo{pages}{387--405}.
\bibitem[{Wang et~al.(2023)Wang, Demir, Mohammad, Oren, and Anantram}]{wang2023computational}
\bibinfo{author}{Y.~Wang}, \bibinfo{author}{B.~Demir}, \bibinfo{author}{H.~Mohammad}, \bibinfo{author}{E.~E. Oren}, \bibinfo{author}{M.~Anantram},
\newblock \bibinfo{title}{Computational study of the role of counterions and solvent dielectric in determining the conductance of b-dna},
\newblock \bibinfo{journal}{Physical Review E} \bibinfo{volume}{107} (\bibinfo{year}{2023}) \bibinfo{pages}{044404}.
\bibitem[{Luo et~al.(2023)Luo, Xu, Peng, Wang, Duan, Mahmood, Wen, Ding, and Xu}]{luo2023aq2pnn}
\bibinfo{author}{Y.~Luo}, \bibinfo{author}{N.~Xu}, \bibinfo{author}{H.~Peng}, \bibinfo{author}{C.~Wang}, \bibinfo{author}{S.~Duan}, \bibinfo{author}{K.~Mahmood}, \bibinfo{author}{W.~Wen}, \bibinfo{author}{C.~Ding}, \bibinfo{author}{X.~Xu},
\newblock \bibinfo{title}{Aq2pnn: Enabling two-party privacy-preserving deep neural network inference with adaptive quantization},
\newblock in: \bibinfo{booktitle}{2023 56th IEEE/ACM International Symposium on Microarchitecture (MICRO)}, \bibinfo{organization}{IEEE}, \bibinfo{year}{2023}, pp. \bibinfo{pages}{628--640}.

\end{thebibliography}


\end{CJK}
\end{sloppypar}
\end{document}